\documentclass[mnsc,nonblindrev]{informs4}

\usepackage{eqndefns-left}    
\RequirePackage{tgtermes}
\RequirePackage{newtxtext}
\RequirePackage{newtxmath}
\RequirePackage{bm}
\RequirePackage{endnotes}     

\OneAndAHalfSpacedXI          

\usepackage{xcolor}
\usepackage{longtable}
\usepackage{enumitem}
\usepackage{multirow}
\usepackage{booktabs}
\usepackage{threeparttable}
\usepackage{pdflscape}
\usepackage{rotating}
\usepackage{subcaption}
\usepackage{float}
\usepackage{adjustbox}
\usepackage{algorithm}
\usepackage{algpseudocode}

\newcommand{\sym}[1]{\ensuremath{^{#1}}}

\newif\ifshowcomments
\showcommentsfalse
\newcommand{\ra}[1]{}
\newcommand{\jl}[1]{}
\newcommand{\rahigh}[1]{}

\let\footnote=\endnote

\usepackage{natbib}
\bibpunct[, ]{(}{)}{,}{a}{}{,}%
\def\bibfont{\small}%

\TheoremsNumberedThrough
\EquationsNumberedThrough
\ECRepeatTheorems

\MANUSCRIPTNO{}

\def\theARTICLETOPLEFT{}    
\def\theARTICLETOPRIGHT{}   
\RRHSecondLine{}            
\LRHSecondLine{}            

\makeatletter
\def\theARTICLEABSTRACT{%
  \HOOKb
  \vspace*{18pt}%
  \begin{minipage}[t]{\textwidth}\parindent1em\ABSfont
    \noindent\theABSTRACT\endgraf
    \vskip5pt
    \theFUNDING
    \theKEYWORDS
    \theSUBJECTCLASS
    \theAREAOFREVIEW
    \theMSCCLASS
    \theORMSCLASS
    \noindent\hrulefill
  \end{minipage}%
  \vspace*{0pt}%
}
\makeatother

\begin{document}

\RUNAUTHOR{Luo, Agarwal, and Gao}

\RUNTITLE{From Testing to Learning}

\TITLE{Beyond One-shot: AI Agents for Learning in Field Experiments}

\ARTICLEAUTHORS{%
\AUTHOR{Junjie Luo}
\AFF{Johns Hopkins School of Medicine, Baltimore, MD, USA, \EMAIL{jluo41@jhu.edu}, \URL{}}
\AUTHOR{Ritu Agarwal}
\AFF{Carey Business School, Johns Hopkins University, Baltimore, MD, USA, \EMAIL{ritu.agarwal@jhu.edu}, \URL{}}
\AUTHOR{Gordon Gao}
\AFF{Carey Business School, Johns Hopkins University, Baltimore, MD, USA, \EMAIL{gordon.gao@jhu.edu}, \URL{}}
}

\ABSTRACT{%
%
Organizations routinely run experiments for A/B testing, yet the data generated from one experiment is underutilized to inform subsequent intervention design.
%
Significant barriers exist to extracting actionable knowledge from prior experimental data to inform new interventions.
%
We study whether tool-augmented agentic AI can automatically learn from experimental data to generate new interventions in subsequent experiments.
%
Through two-stage field experiments in healthcare prescription messaging (693,139 patient visits), we compare a Human + Chatbot method (Stage~1: behavioral experts with conversational AI co-designing 13 message variants, 444,691 patient visits) against a Tool-Augmented Agentic AI method (Stage~2: AI autonomously extracting principles from Stage~1 data to generate 17 new variants, 248,448 patient visits).
%
The Agentic AI method, equipped with analytical tools, structured Data-Information-Knowledge-Wisdom (DIKW) reasoning agents, and transparent evidence chains, produces superior interventions: the best AI-generated message achieved a 69.8\% CTR (+6.5 percentage points over baseline).
%
Critically, our results suggest that the value comes from domain-specific experimental data, not from general reasoning ability: frontier LLMs operating without experimental data failed to predict which interventions would succeed.
%
The field experiments also revealed that general-purpose behavioral theories used for intervention design do not extend uniformly to specific healthcare contexts, motivating an agentic AI approach to theory audits at field-experiment scale.
%
Our research shows that tool-augmented AI can learn from experimental data and generate improved domain-relevant interventions, transforming behavioral experimentation from one-shot evaluation into a scalable system for cumulative design learning.
}%



\maketitle

\section{Introduction}

%
One of the most important developments in behavioral economics has been the extensive use of field experimentation beyond the laboratory into real-world markets, firms, and policy settings \citep{list2024field}.
%
Indeed, behavioral experiments have become a standard tool for evidence-based decision making, with organizations increasingly running megastudies for greater comparability across different treatments \citep{milkman2021megastudies}.
%
%
In healthcare, these experiments are the gold standard for evaluating behavioral interventions across a variety of settings such as medication adherence programs, vaccination campaigns, and chronic disease management \citep{thakkar2016mobile,sun2019mobile, milkman2024megastudy,schillinger2021precision}.
%
Such experiments can be large: a single healthcare messaging study may involve hundreds of thousands of patient interactions across multiple treatment variants and demographic subgroups \citep{milkman2024megastudy,schillinger2021precision}.
%

Ideally, the corpus of rich experimental data from such studies would translate into cumulative design knowledge that informs superior future interventions.
%
Yet to date, most interventions are one-shot in nature: candidate designs are tested in parallel to identify the best performer \citep{duckworth2022guide,paley2023crowdsourcing,saccardo2024field}, with no second round informed by what was learned.
%
The fact that the experiment data seldom converts into reusable design knowledge has been criticized as hampering the advance of social science \citep{almaatouq2024beyond}.
%
As has been observed, the current paradigm is inadequate; integrative, multi-shot experiments should accumulate insights and guide the next experiment.
%
Potentially valuable evidence about why an intervention worked, for whom it worked, and how its components might be recombined is left unused in the next design cycle.
%
Each experiment thus becomes an isolated evaluation exercise rather than a contribution to cumulative experimental learning.
%

%
One reason for the above ``one-shot'' conundrum is the complexity of extracting actionable design knowledge from experimental data.
%
It requires interpreting statistical patterns through behavioral logic and translating those insights into new intervention designs.
This combination of analysis, interpretation, and synthesis is costly in time, expertise, and organizational attention \citep{kahneman2011thinking}.
%
Solving this problem requires three capabilities that existing approaches rarely combine: (1) working systematically with large-scale experimental data, (2) reasoning across multiple levels of abstraction, from statistical patterns to behavioral mechanisms to design principles, and (3) maintaining grounding in domain-specific evidence rather than general-purpose cross-domain heuristics \citep{almaatouq2024beyond,kohavi2020trustworthy}.
%
Expert analysis provides behavioral interpretation but cannot systematically explore megastudy-scale pattern spaces \citep{kahneman2009conditions}.
%
Statistical software provides computational scale but stops at coefficients and p-values rather than synthesizing behavioral mechanisms or design principles.
%
Frontier language models generate plausible content from generalized training data but have no access to an organization's own experimental results and no analytical tools to process them \citep{ji2023survey,lewis2020retrieval}.
%
No existing approach combines all three capabilities, hindering the ability of researchers to sustain cumulative learning from their experiments.
%

We study whether tool-augmented agentic AI can serve as a cumulative knowledge system that learns from experimental data to generate superior interventions in subsequent experiments.
%
Our agentic system uses structured DIKW reasoning to progress from raw experimental data through statistical patterns and behavioral mechanisms to design principles and next-round interventions \citep{ackoff1989data,rowley2007wisdom}.
%
Critically, the value lies not in the LLM's general reasoning ability but in the learning infrastructure: code execution for statistical analysis, explicit evidence chains linking each reasoning level back to specific experimental observations, and structured progression across abstraction levels.
%
A key property of this infrastructure is transparency and explainability: every design decision can be traced back through the reasoning levels to specific experimental observations.
%
The tools provide computational scale and evidence grounding; AI provides the capacity to move from statistical patterns to behavioral principles; neither is sufficient alone, but their combination addresses all three capabilities identified above.
%
The result is a cumulative learning process in which one experiment can inform the design of the next.
%

We investigate three research questions related to whether and how cumulative experimental learning can be realized.
%
First, regarding this novel approach of agentic AI-based cumulative experimental learning, we ask whether the experimental data from a megastudy can be transformed into reusable design knowledge that informs the next round of interventions.
%
Second, we investigate how tool-augmented AI systems should be designed to perform this learning: which architectural components and design principles enable the transformation of statistical patterns into actionable behavioral knowledge.
%
Third, we test if interventions generated by such a system outperform prior-round designs and frontier LLMs operating without experimental data.
%

\begin{figure}[t]
    \centering
    \includegraphics[width=\textwidth]{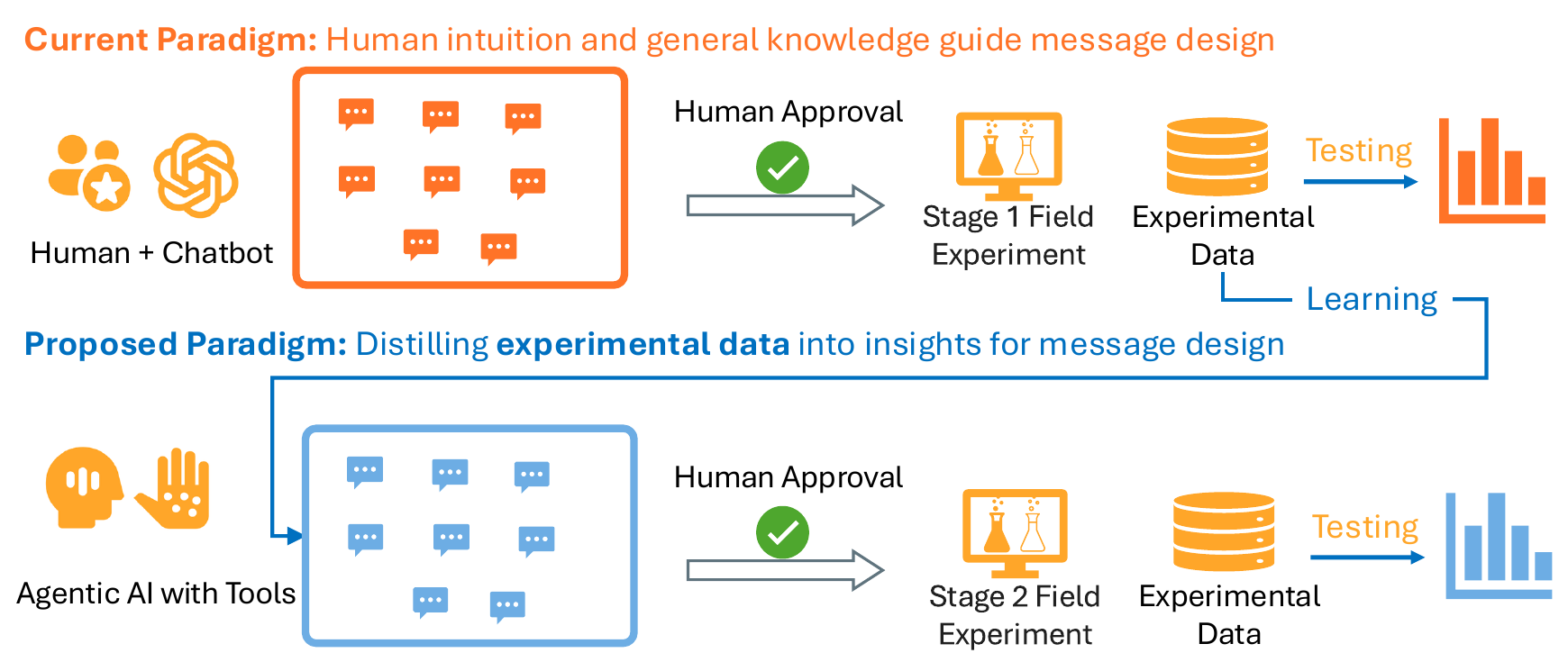}
    \caption{Conceptual Overview of Two-Stage Learning Architecture. Stage~1 tests multiple intervention variants through a randomized field experiment. Stage~2 uses tool-augmented agentic AI to extract behavioral principles from Stage~1 data and generate new interventions, which are then validated in a second field experiment.}
    \label{fig:intro-workflow}
\end{figure}

Our research setting is patient prescription messaging.
%
To address the above research questions, we conducted two-stage field experiments, which involved a total of 693,139 patient visits (Figure~\ref{fig:intro-workflow}).
%
In Stage~1, behavioral experts working with GPT-4 co-designed 13 message variants that were tested on 444,691 patient visits, generating the experimental data used for subsequent knowledge extraction.

%
In Stage~2, we designed and deployed tool-augmented agentic AI that uses code execution, statistical analysis, and structured DIKW reasoning to extract behavioral principles from the Stage~1 data and generate 17 new message variants, which were tested in a randomized trial in 248,448 patient visits.
%
Previewing our results, the best AI-generated message achieved a 69.8\% click-through rate, 6.5 percentage points above the Stage~2 baseline (a 10.3\% relative improvement), while frontier LLMs operating without experimental data failed to predict which interventions would succeed.
%
Efficiency framing and professional authority emerged as reliable principles in this healthcare setting, while social proof and reciprocity proved ineffective, surfacing domain-specific behavioral knowledge that general nudging theory does not anticipate.
%
Together, these results show that behavioral experiments can function as cumulative knowledge systems when paired with tool-augmented infrastructure, and that domain-specific experimental data, not model capability alone, drives the learning cycle.

\section{Literature Review}

%
Four streams of literature provide the conceptual grounding for our work: cumulative learning from behavioral experiments (Section~2.1), tool-augmented agentic AI (Section~2.2), knowledge hierarchies and design science (Section~2.3), and healthcare messaging as a test domain (Section~2.4).

\subsection{Learning from Behavioral Experiments}

%
As discussed, organizations rarely sustain cumulative learning from behavioral experiments, defaulting instead to one-shot evaluation \citep{milkman2021megastudies,duckworth2022guide,paley2023crowdsourcing,feng2022randomized,ghose2024effects,li2021peer}.
%
\citet{almaatouq2024beyond} diagnose this pattern more broadly, arguing that social science experimentation amounts to ``playing 20 questions with nature,'' where integration of findings across studies is assumed to happen through the publication process but rarely does.
%
Organizational learning theory signifies the importance of learning from past experience: without explicit processes for encoding and retrieving lessons, experiential knowledge remains tacit and organizations default to exploiting known winners rather than investing in deeper extraction \citep{levitt1988organizational,march1991exploration,argote2012organizational}.
%
Knowledge management systems address the storage and retrieval of existing expertise \citep{alavi1999knowledge} but not the creation of new knowledge from raw experimental data \citep{nonaka1994dynamic}.
%
What is missing is knowledge creation infrastructure that transforms experimental results into reusable design principles, enabling cumulative learning across experimental cycles.
%
Within this literature, we ask whether the experimental data from one round of behavioral experiments can be transformed into reusable design knowledge for the next round.
%
In this paper, we designed and field-tested a tool-augmented agentic AI system that uses structured DIKW (Data-Information-Knowledge-Wisdom) reasoning to extract behavioral principles from prior experimental data and generate new interventions for subsequent experimentation.
%
Our paper contributes by showing that tool-augmented agentic AI can bridge the long-standing gap between experimental results and reusable design knowledge, opening a path from one-shot evaluation to cumulative experimental learning.

\subsection{From LLMs to Tool-Augmented Agentic AI}

%
Large language models have demonstrated value in creative and knowledge work, improving performance in a variety of domains, including ad copywriting \citep{chen2024large}, management consulting \citep{dell2023navigating}, design work \citep{hou2025double}, and professional writing \citep{noy2023experimental,brynjolfsson2025generative}.
%
However, the modality of human-AI collaboration matters: \citet{chen2024large} find that using LLMs as ghostwriters can anchor users and reduce output quality, while using them as sounding boards helps non-experts close the gap with experts.
%
More fundamentally, LLMs operating from general training data produce plausible but ungrounded content when applied to domain-specific tasks \citep{ji2023survey}, and show poor calibration in specialized contexts such as clinical decision-making \citep{shool2025systematic}.
%
The ``AI as prediction machine'' framing \citep{agrawal2022prediction} captures AI's value in well-defined prediction tasks but does not address knowledge creation from organizational experimental data.
%
The key limitation is that without access to an organization's own experimental results and without analytical tools to process them, LLMs can only generate from general patterns rather than extract from specific evidence.
%

Tool-augmented language models address part of this limitation by enabling LLMs to execute code, run statistical analyses, and interact with external data sources \citep{schick2023toolformer,patil2024gorilla}.
%
Such capabilities go beyond retrieval-augmented generation (RAG), which retrieves relevant text passages to condition LLM outputs \citep{lewis2020retrieval}: here, the system executes computational analyses on data and reasons over the results, rather than merely retrieving and summarizing existing text.
%
This shifts the task from content generation, i.e., producing plausible text from training-data patterns, to evidence-based extraction: analyzing specific datasets and deriving grounded conclusions.
%
Because the system executes code to produce statistical results before the LLM interprets them, the reasoning is predicated on verified computational outputs rather than training-data associations.
%
Multi-agent architectures extend this further by coordinating specialized components, each operating at a distinct level of abstraction, enabling structured progression from raw data through statistical patterns to behavioral mechanisms and design principles \citep{park2023generative,wu2024autogen}.
%
These capabilities have been applied to software engineering \citep{jimenez2023swe}, scientific research \citep{lu2024ai}, and data analysis \citep{hong2025data}, though primarily in benchmark evaluations rather than real-world organizational settings; cumulative learning from behavioral experiments remains untested.
%
Within this design space, we ask how multi-agent systems should be architected to close the experimental learning loop that remains an open question in the current literature.
%
Our paper contributes to this literature by showing that domain-specific experimental data, analytical tools, and structured multi-level reasoning can transform generative AI from a content-generation chatbot into a tool-augmented agentic AI system that performs evidence-based knowledge extraction.

\subsection{Knowledge Hierarchies and Design Science}

%
The Data-Information-Knowledge-Wisdom (DIKW) hierarchy \citep{ackoff1989data,rowley2007wisdom,jennex2009re} proposes that raw data gains value through progressive abstraction: data becomes information through contextualization, information becomes knowledge through interpretation and pattern recognition, and knowledge becomes wisdom through application to novel decision contexts.
%
The hierarchy maps naturally onto the experimental learning problem: raw experimental results (data) must be processed into statistical patterns (information), interpreted as behavioral principles (knowledge), and applied to generate new intervention designs (wisdom).
%
This multi-level structure explains why no single existing approach, whether expert analysis, statistical software, or general-purpose LLMs, spans the full hierarchy.
%
We approach this problem from a design science perspective \citep{hevner2004design,gregor2013positioning}, building and evaluating an artifact that instantiates these DIKW transitions through tool-augmented AI with specialized agents operating at each level.
%
Our paper contributes to the design science literature by operationalizing the DIKW hierarchy as a working multi-agent artifact.
%
The architecture renders each level-to-level transition auditable, with transparent evidence chains linking every design decision back to specific experimental observations.

\subsection{Healthcare Messaging as a Learning Domain}

%
In healthcare settings, messaging has emerged as a widely used trigger for changing patient behavior across vaccination campaigns, chronic disease management, and preventive care. 
Our focus in the study is on a critical challenge in healthcare: prescription medication adherence.
%
Medication nonadherence affects approximately 50\% of patients with chronic conditions \citep{brown2011medication}, and the broader problem of non-optimized medication therapy generates an estimated \$528 billion in annual US healthcare costs \citep{watanabe2018cost}.
%
Digital health interventions, particularly SMS and email messaging, have emerged as scalable approaches for improving adherence and patient engagement \citep{thakkar2016mobile,sun2019mobile,ghose2022empowering,zhou2022turn}.

Yet effectiveness varies sharply with design details such as sender identity, framing, action specificity, and timing conditioning the effects of the intervention \citep{schillinger2021precision,so2017message,li2024impact}.
%
Even well-designed trials struggle: a recent large-scale pragmatic study of nudge messages for cardiovascular medication adherence found no significant improvement in refill rates \citep{ho2025personalized}.
%
The messages in our study function as digital nudges \citep{thaler2009nudge}; we focus, however, not on nudging theory itself but on the upstream organizational challenge of learning cumulatively from nudging experiments.

%
We note that the healthcare context has distinct characteristics and patient communication represents a fundamentally different context from consumer messaging.
%
Privacy and autonomy dominate patient decision-making rather than brand engagement \citep{kwame2021literature,adjerid2018beyond}, and professional authority anchors credibility in ways that commercial endorsement cannot.
%
Patient populations also vary widely in health literacy \citep{chang2017leveraging} and technology adoption \citep{czaja2006factors}, producing heterogeneity that makes the transfer of consumer-context insights to healthcare uncertain.
%
In prior work, methodologically adjacent approaches have used multi-armed bandit algorithms to dynamically personalize across existing healthcare interventions \citep{zhou2023spoiled}.
%
Our approach differs from this in that we instead extract transferable design principles from completed experiments to inform the design of new interventions for the next round.
%
Within this domain, we ask whether interventions generated by tool-augmented agentic AI outperform expert-designed baselines and whether the resulting principles surface domain-specific behavioral knowledge that general nudging theory does not anticipate.
%
Our paper contributes to the healthcare-messaging and behavioral-nudging literatures by demonstrating that data-driven principle extraction can recover effective interventions in a domain where general nudging theory has shown limited transfer.
\section{Research Design}

\subsection{Empirical Setting}

\begin{table}[t]
\centering
\caption{Dataset Overview and Study Variables}
\label{table:dataset-overview}
\footnotesize
\begin{tabular}{p{3.8cm} p{4.8cm} c c}
\toprule
 & & \textbf{Stage~1} & \textbf{Stage~2} \\
\midrule
\multicolumn{4}{l}{\textit{\textbf{Study Design}}} \\
\quad Study period   & Dates of experiment     & Jun 16--Jul 3, 2025  & Aug 25--Sep 8, 2025 \\
\quad Duration       & Days of enrollment      & 18 days              & 15 days             \\
\quad Message variants & Number tested         & 13                   & 20$^\dagger$        \\
\quad Design method  & How messages were created & Human + LLM        & Agentic AI (DIKW)   \\
\quad Invitations    & Total patient visits    & 444,691              & 248,448             \\
\addlinespace
\multicolumn{4}{l}{\textit{\textbf{Patient Variables}}} \\
\quad Age 18--44     & Younger age cohort (\%) & 48.9                 & 50.0                \\
\quad Age 45--64     & Older age cohort (\%)   & 51.1                 & 50.0                \\
\quad Female         & Female patient (\%)     & 64.5                 & 65.3                \\
\quad Location       & State + Area Deprivation Index (ADI) & 50 states & 50 states \\
\addlinespace
\multicolumn{4}{l}{\textit{\textbf{Medical Context Variables}}} \\
\quad Drug category  & Acute / chronic / mental health & multi-category & multi-category \\
\quad Therapeutic category & Drug class (e.g., cardiovascular, psychiatric) & multi-category & multi-category \\
\quad Provider specialty & Prescriber specialty (primary care, specialist, etc.) & multi-category & multi-category \\
\quad Message variant & Randomly assigned SMS text & 13 variants & 20 variants \\
\addlinespace
\multicolumn{4}{l}{\textit{\textbf{Outcome Variables}}} \\
\quad Click-through  & \textit{Primary}: patient clicked link (\%) & 59.4 & 62.6 \\
\quad Authentication & \textit{Secondary}: identity verified (\%) & 46.9 & 50.0 \\
\midrule
\textbf{Total} & \multicolumn{3}{l}{693,139 invitations across 2 stages} \\
\bottomrule
\end{tabular}

\vspace{0.3em}
\begin{minipage}{\textwidth}
\scriptsize
$^\dagger$ Stage~2: 17 AI-generated messages + 3 Stage~1 baselines (\textit{salience}, \textit{progressFeedback}, \textit{default}).
Randomization balance confirmed ($\chi^2$ $p > 0.05$, both stages; see Appendix~\ref{app:experimental-validation}).
\end{minipage}

\end{table}

%
We evaluate our approach in the context of prescription notification messaging, which, as noted earlier, is a healthcare domain where systematic experimental learning is particularly valuable (Table~\ref{table:dataset-overview}).
%
For our field experiments, we collaborated with one of the largest patient messaging platforms in the United States, which processes millions of prescription notifications annually across thousands of healthcare providers. 
%
We focus on the prescription notification aspect of their portfolio. This provides access to large-scale experimental data on prescription notification interventions.

\noindent\textbf{The Intervention}
%
When a healthcare provider writes a new prescription, the platform sends a text message (SMS) to the patient notifying them that prescription information is available for review.
%
The message includes a brief greeting attributed to the provider's office, a short description of the action, and a link to a secure portal where the patient can view prescription details.
%
A typical message reads: ``Hi, it's Dr.\ Johnson's office.\ Review your Rx details here: [link].''
%
The patient's response is a binary choice: click the link to engage, or ignore the message.
%
If the patient clicks the link, they may then authenticate their identity and review the prescription.
%
The only element that varies across experimental conditions is the message text itself; all other elements (sender, link, timing infrastructure) remain constant.

\noindent\textbf{Dataset Structure}
%
Table~\ref{table:dataset-overview} provides an overview of the two field experiments: Stage~1 (June~16--July~3, 2025; 444,691 patient visits across 13 human-and-LLM co-designed message variants) and Stage~2 (August~25--September~8, 2025; 248,448 patient visits across 20 variants generated by the tool-augmented agentic AI system).
%
The table also summarizes the study variables we use throughout, the randomly assigned message variant, and engagement outcomes.
%
Each patient visit record contains contextual information in four categories.
%
\textit{Patient demographics} include age (18--44, 45--64), gender, and geographic location (state, area deprivation index).
%
\textit{Medical context} captures prescription details (drug name, therapeutic category), provider characteristics (specialty, credentials), and clinical metadata.
%
\textit{Temporal factors} document message delivery and response timing.
%
\textit{Experimental assignment} records the randomly assigned message variant with its text and character length.

\noindent\textbf{Outcome Measures}
%
We employ a hierarchical outcome structure reflecting progressive patient engagement.
%
The \textit{primary outcome} is message click-through (binary), indicating whether the patient clicked the link in the SMS.
%
The \textit{secondary outcome} is authentication, indicating whether the patient completed identity verification after clicking the link, which represents a deeper level of engagement than click-through alone.
%
We focus here primarily on click-through rates as the most proximal and reliable measure of message effectiveness.

\subsection{Stage~1: Message Design and Field Experiment}


%
Stage~1 represents the current state of practice for AI-assisted intervention design \citep{chen2024large,noy2023experimental}: behavioral science researchers collaborating with a conversational LLM to generate theory-driven message variants.
In OpenAI's framework of AI capabilities,\footnote{OpenAI defines five levels of AI: Level~1 (Chatbots: conversational language), Level~2 (Reasoners: human-level problem solving), Level~3 (Agents: systems that take actions), Level~4 (Innovators: aid in invention), Level~5 (Organizations: do organizational work). 
See \citet{bloomberg2024openai}.} this represents Level~1 AI (Chatbots): conversational AI that generates content based on prompts but without autonomous reasoning, tool use, or systematic data analysis capabilities.

%
\noindent\textbf{Stage 1 Message Design Process}
%
Six researchers with expertise in behavioral science, health communication, and experimental design, developed the Stage~1 message portfolio through a structured four-step process.
%
First, the team selected psychological principles through group deliberation, drawing on established behavioral science frameworks including Cialdini's principles of influence (authority, social proof) \citep{cialdini2009influence,cialdini2007descriptive}, prospect theory (gain-loss framing) \citep{kahneman2013prospect}, and health communication research (temporal salience, commitment devices, progress feedback, emotional appeals, goal reinforcement).
%
Second, the lead researcher generated three candidate messages per principle using GPT-4, which was the leading commercial conversational LLM at the time of Stage~1 (June 2025).
%
The model was prompted with the principle definition, the prescription notification context, and structural constraints (maximum 84 characters for message text, provider attribution required, no false urgency claims).
%
Third, the full research team evaluated all 39 candidates and selected one message per principle by majority vote, yielding 13 final message variants.
%
Finally, the messaging platform company's technology leadership team, independent of the research team, reviewed all 13 messages for clinical appropriateness and regulatory compliance prior to deployment.

%
This design process drew on cross-domain behavioral theory and the LLM's general training data rather than systematic analysis of healthcare-specific experimental evidence, because no prior experimental data existed in this notification context.
%
Stage~1 therefore serves a dual role: testing a diverse portfolio of theory-driven interventions and generating the experimental dataset that is used in Stage~2 for knowledge extraction.

%
\noindent\textbf{Stage 1 Field Experiment}
%
The 13 messages were deployed in a randomized controlled trial with 444,691 patient visits over 18 days (June 16 to July 3, 2025).
%
Each visit was randomly assigned to one of the 13 variants; messages also varied in linguistic features including greeting style, call-to-action placement, and character length (full portfolio in Appendix~\ref{app:stage1-messages}).
%
The resulting experimental dataset becomes the input for the DIKW learning process described next.

\subsection{DIKW Learning Process}

\begin{figure}[t]
    \centering
    \includegraphics[width=\textwidth]{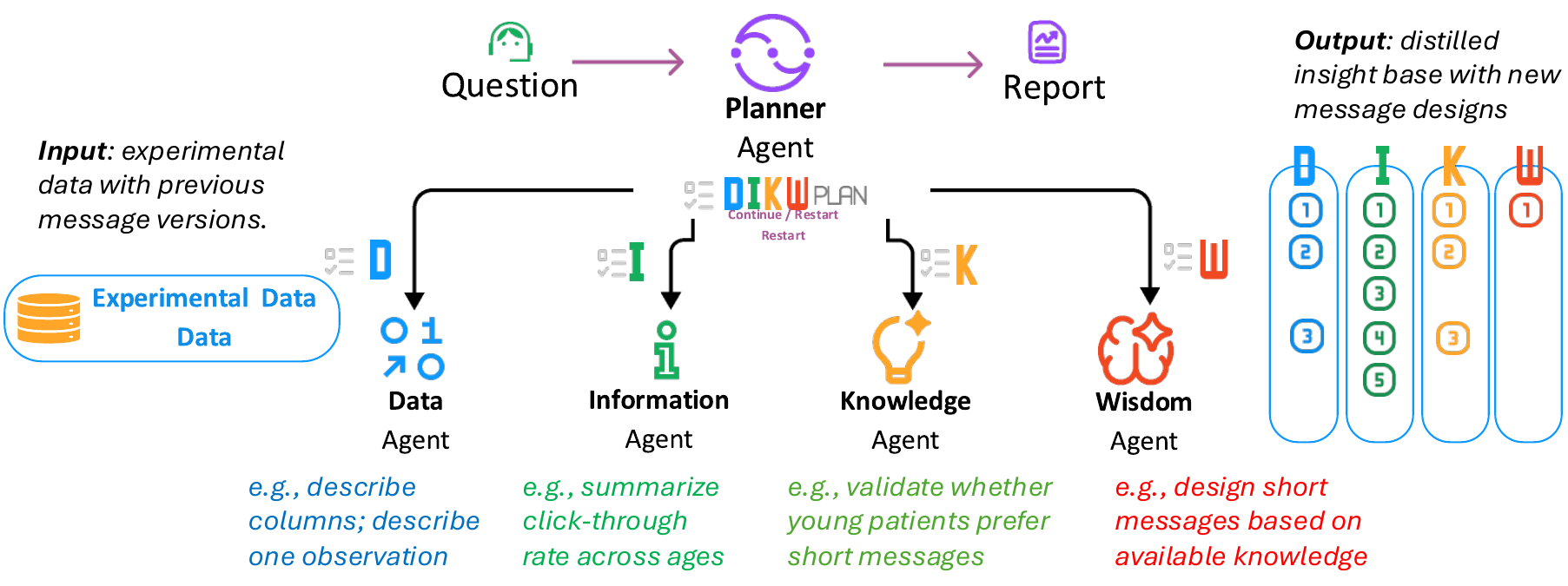}
    \caption{DIKW Multi-Agent System Architecture. An orchestrator agent determines analytical topics, which are assigned to specialized agents at four levels: the D-Agent validates data integrity, the I-Agent extracts statistical patterns through code execution, the K-Agent synthesizes findings into behavioral principles, and the W-Agent translates principles into intervention designs. The human researcher reviews and approves the plan before execution begins.}
    \label{fig:dikw-workflow}
\end{figure}

%
As noted, experimental learning is accomplished through a multi-agent system constructed on the DIKW framework (Figure~\ref{fig:dikw-workflow}; formal agent specifications and computational architecture in Appendix~\ref{app:dikw-technical}). 
%
As depicted in Figure~\ref{fig:dikw-workflow}, the system operates through an orchestrator agent and four specialized agents (the D-Agent, I-Agent, K-Agent, and W-Agent) with two distinct execution modes.
%
An orchestrator agent first generates a plan of analytical topics at each DIKW level: Data-level topics covering dataset validation and structure, Information-level topics covering engagement metrics, demographic patterns, temporal dynamics, and cross-dataset consistency, Knowledge-level topics address hypothesis testing and principle synthesis, and a Wisdom-level topic is used for message generation.
%
The human researcher reviews and approves this plan before execution begins.
%
All agents were powered by Claude-4 (Sonnet) \citep{anthropic2025claude4} with layer-specific prompting strategies (Appendix~\ref{app:agent-prompts}).
%
Agent orchestration, inter-agent message passing, and persistent state across plan sessions were implemented using LangGraph \citep{langgraph2025}, an open-source framework for building stateful multi-agent LLM applications.

%
Formally, each level is realized by an Agent-Unit that maps a task specification together with lower-level artifacts to a structured output:
\begin{align*}
\mathcal{D} &: \big(D,\, \mathcal{T}^{(D)}_i.\text{spec}\big) \\
&\quad \longrightarrow \mathcal{T}^{(D)}_i.\text{output}, \\
\mathcal{I} &: \big(D,\, \{\mathcal{T}^{(D)}_i.\text{output}\}_i,\, \mathcal{T}^{(I)}_j.\text{spec}\big) \\
&\quad \longrightarrow \mathcal{T}^{(I)}_j.\text{output}, \\
\mathcal{K} &: \big(\{\mathcal{T}^{(I)}_j.\text{output}\}_j,\, \mathcal{T}^{(K)}_k.\text{spec}\big) \\
&\quad \longrightarrow \mathcal{T}^{(K)}_k.\text{output}, \\
\mathcal{W} &: \big(\{\mathcal{T}^{(K)}_k.\text{output}\}_k,\, K^{\text{open}},\, \mathcal{T}^{(W)}_\ell.\text{spec}\big) \\
&\quad \longrightarrow \mathcal{T}^{(W)}_\ell.\text{output}.
\end{align*}
%
Here $\mathcal{T}^{(L)}_i.\text{spec}$ encodes the task and $\mathcal{T}^{(L)}_i.\text{output}$ the produced artifacts; sets such as $\{\mathcal{T}^{(D)}_i.\text{output}\}_i$ denote all available lower-level outputs, and $K^{\text{open}}$ is the open-domain prior knowledge invoked at the Wisdom level.
%
Full output-tuple definitions and the bidirectional plan-revision protocol are provided in Appendix~\ref{app:dikw-technical}.

%
At the Data and Information levels, the D-Agent and I-Agent operate using autonomous code generation and execution.
%
For each topic, a specialized agent receives a task specification and the experimental dataset, then autonomously writes Python code (typically 150 to 300 lines using pandas, scipy, and matplotlib) to perform the required analysis.
%
The code executes in a sandboxed environment, producing three artifacts: the executable script itself (preserved for reproducibility), statistical tables in CSV format, and a structured report summarizing findings with embedded figures.
%
At the Data level, validation tasks such as assessing data quality and completeness, verifying randomization balance across treatment arms, and documenting the experimental design parameters are performed.
%
At the Information level, topics become analytical, each producing individual statistical findings about the Stage~1 data: one topic computed click-through rates stratified by age group, gender, income level, and geographic region; another analyzed how engagement varied across psychological framing strategies; a third identified temporal patterns in message response timing.

%
At the Knowledge and Wisdom levels, the K-Agent and W-Agent shift from code generation to reasoning synthesis.
%
This separation is architecturally critical: because the K-Agent reads from verified I-level computational outputs rather than generating from training data, its knowledge assessments are grounded in reproducible statistical evidence rather than plausible-but-unverified associations. The system cannot claim a behavioral pattern exists unless I-level code execution has already produced numerical evidence for it.
%
Each agent reads the statistical reports produced at lower levels, selects the most relevant findings, evaluates the strength and generalizability of the evidence, and produces a reasoning analysis.
%
At the Knowledge level, the K-Agent synthesizes multiple Information-level findings into integrated assessments of which messaging principles are reliable in the domain; no single statistical finding establishes a principle on its own, so the K-Agent combines converging evidence across analyses, with explicit evidence chains back to the underlying Information-level reports.
%
A key dimension of this evidence synthesis is cross-subgroup consistency: the K-Agent's confidence scores explicitly weight whether a pattern holds across medical contexts (acute versus chronic, mental health versus physical), demographic subgroups, and temporal conditions: principles that are robust across subgroups receive higher confidence than those that hold only in aggregate.
%
For instance, one topic tested whether demographic-specific engagement patterns observed in Stage~1 held consistently across age, gender, and medical-context subgroups.
%
When a Knowledge-level task determines that available evidence is insufficient, the system can pause and request additional Data or Information analyses, implementing the bidirectional flow that distinguishes this from a simple sequential pipeline.

%
At the Wisdom level, we asked the agent to generate 20 candidate message designs for Stage~2 testing, the maximum number of variants the messaging platform permits in a single experiment.
%
Generation operated under the same operational constraints as Stage~1: an 84-character cap on the message body (the residual budget within the 160-character SMS after the prescription link and opt-out text), required provider attribution, and a prohibition on misleading claims about the linked content.
%
Each design was accompanied by a rationale tracing back to specific Knowledge-level assessments and their supporting Information-level evidence.
%

The 20 messages generated by the system for Stage~2 testing underwent medical, legal, and operational review by the messaging platform's technology leadership team (the same independent team that reviewed Stage~1 messages) for clinical appropriateness, regulatory compliance, and operational feasibility.
%
Three messages were excluded: \textit{autonomyMax} (too passive for healthcare urgency), \textit{microCommitment} (ambiguous question format for SMS), and \textit{stepCompletionUrgency} (redundant with other completion-framed messages).
%
The remaining 17 AI-generated messages, combined with 3 top-performing Stage~1 baselines (\textit{salience}, \textit{progressFeedback}, \textit{default}), formed the 20-variant portfolio for Stage~2 validation.
%
Each generated message maintains an explicit evidence chain tracing back through Knowledge, Information, and Data levels to specific Stage~1 observations.
%
This ensures that every design decision traces to empirical support rather than speculative extrapolation, enabling \textit{post hoc} verification of the reasoning process.

\subsection{Stage~2: Field Experiment}

\noindent\textbf{Experimental Protocol}
%
Stage~2 tests whether the knowledge extracted and synthesized by the tool-augmented AI agents produces superior interventions, through a randomized controlled trial on the same prescription notification platform used in Stage~1.
%
The experiment ran for 15 days (August 25 to September 8, 2025), approximately seven weeks after the conclusion of Stage~1 (June 16 to July 3, 2025), with no overlap in patient visits between the two stages.
A total of 248,448 patient visits were included in the experiment, each randomly assigned to one of the 20 message variants comprised of 17 AI-generated messages from the DIKW learning process and 3 Stage~1 baselines.
%
The 17 AI-generated messages were designed by the Wisdom-level agent based on the Knowledge-level assessments described in Section~3.3; each message operationalized one or more reliable principles identified from Stage~1 data, and each was accompanied by a design rationale tracing back to specific statistical evidence.
%
The three Stage~1 baselines were selected to provide meaningful comparisons: \textit{default} (the existing standard message, serving as the control), \textit{salience} (the top Stage~1 performer), and \textit{progressFeedback} (the second-best Stage~1 performer).
%
This design allows us to assess whether AI-generated messages outperform both the existing standard and the best available human-designed alternatives (Table~\ref{tab:wisdom-messages} in Section~4.2; full specifications in Appendix~\ref{app:stage2-messages}).

\noindent\textbf{Randomization}
%
Randomization was stratified by patient age category (18--44, 45--64), prescription therapeutic category (acute, chronic, mental health), and provider specialty to ensure balance across key covariates identified as moderators in Stage~1 analysis.
%
With approximately 12,400 visits per variant, the design provides 80\% statistical power to detect a 2.5\% difference in click-through rate between any variant and the default baseline ($\alpha = 0.05$, two-sided).
%
Randomization balance checks are reported in Appendix~\ref{app:experimental-validation}.
%
The experimental protocol and analysis plan were reviewed by the healthcare platform's institutional review board prior to deployment.

\noindent\textbf{Outcome Measurement and Analysis}
%
As noted in Section~3.1, we used click-through rate as the primary outcome, measured identically across both stages to ensure comparability. 
The analysis employs intention-to-treat principles \citep{gupta2011intention}, commonly used in healthcare studies, analyzing all randomized patients regardless of message delivery success or subsequent platform interactions.
%
Statistical comparisons use logistic regression with robust standard errors clustered by provider, adjusting for stratification variables to improve precision.
%
All statistical tests use two-sided $\alpha = 0.05$ with Holm-Bonferroni correction for multiple comparisons.

\section{Results}

\subsection{Stage~1 Results}

\begin{figure}[t]
\centering
\begin{subfigure}[t]{0.48\textwidth}
\centering
\includegraphics[width=\textwidth]{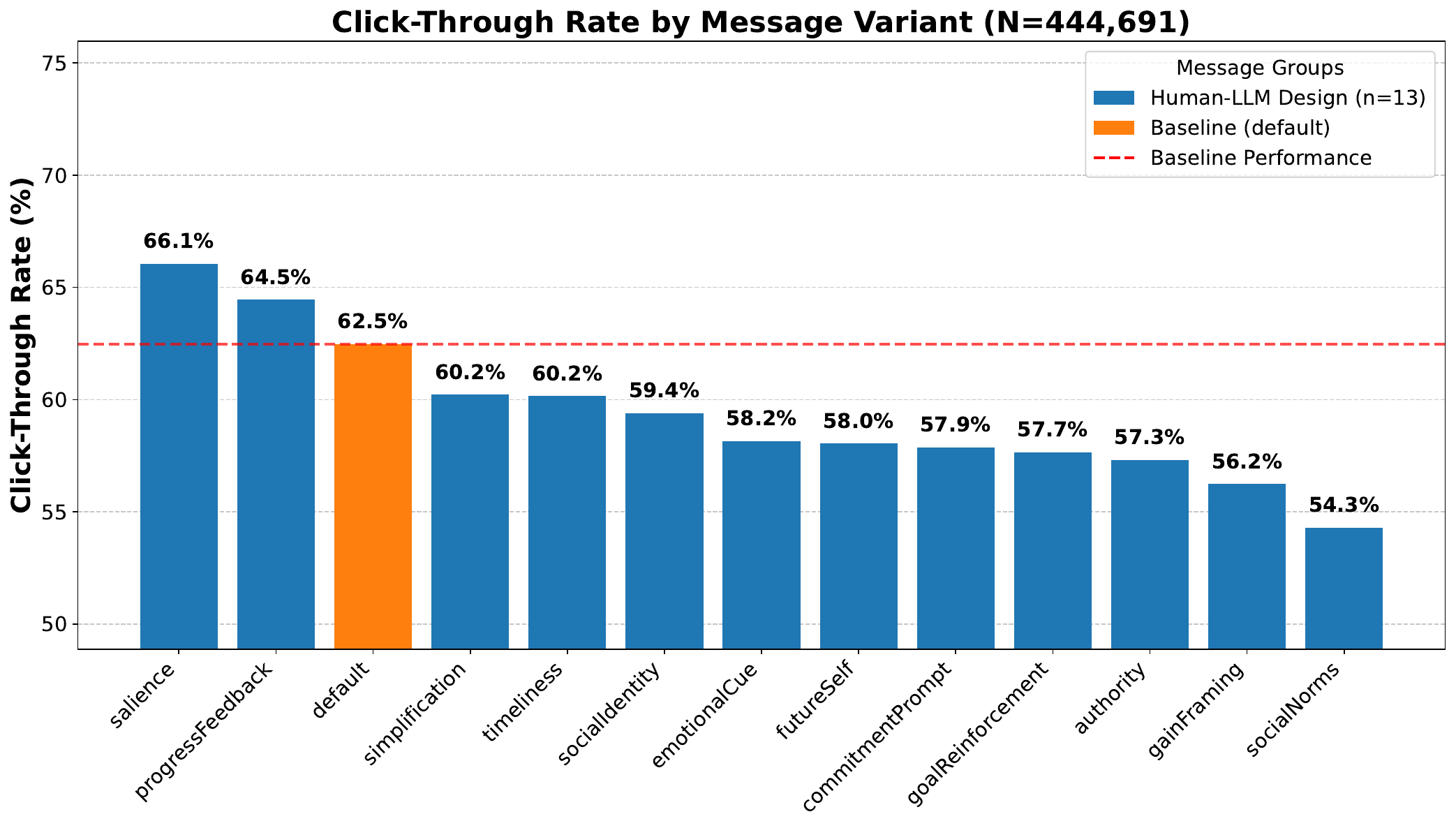}
\caption{Click-Through Rate}
\label{fig:s1-ctr}
\end{subfigure}
\hfill
\begin{subfigure}[t]{0.48\textwidth}
\centering
\includegraphics[width=\textwidth]{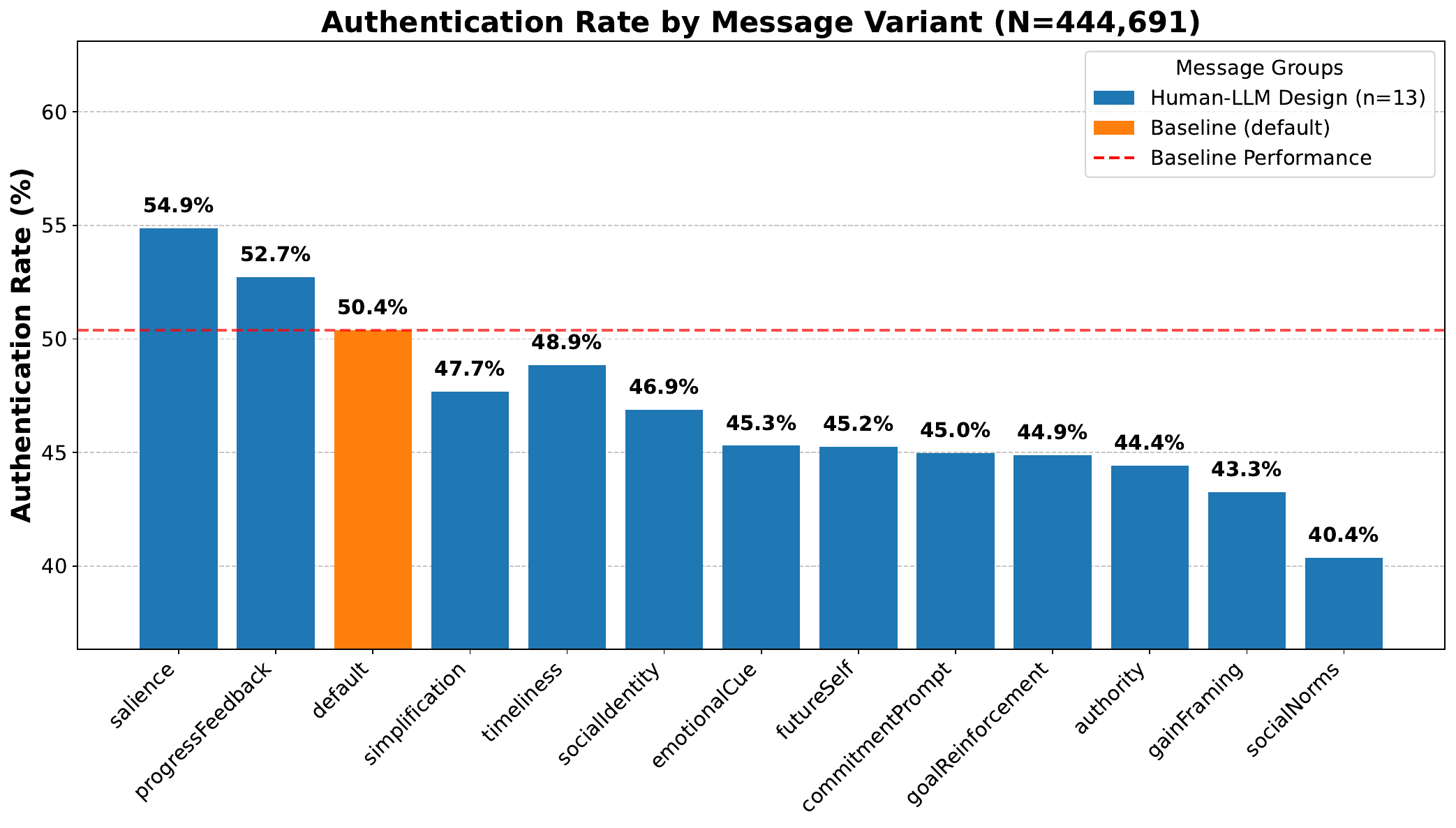}
\caption{Authentication Rate}
\label{fig:s1-auth}
\end{subfigure}
\caption{Stage~1 Message Performance by Variant (N=444,691). (a) Click-through rates for all 13 human-chatbot co-designed messages. (b) Authentication completion rates confirm similar ranking patterns.}
\label{fig:s1-performance}
\end{figure}

%
In the Stage~1 experiment, we tested 13 message variants across 444,691 patient prescription visits (Figure~\ref{fig:s1-performance}).
%
Click-through rates ranged from 54.3\% (socialNorms) to 66.1\% (salience), an 11.8\% spread indicating substantial variation in intervention effectiveness.
%
The top three performers were \textit{salience} (66.1\%), \textit{progressFeedback} (64.5\%), and \textit{default} (62.5\%).
%
Messages emphasizing information salience and task completion outperformed those relying on social proof (socialNorms: 54.3\%), gain framing (gainFraming: 56.2\%), or simple authority (authority: 57.3\%).
%
The authentication panel in Figure~\ref{fig:s1-performance}(b) shows the same ranking, with the top click-through performers also leading on identity verification, indicating the ordering reflects deeper engagement rather than only the initial click.
%
These data, including both the aggregate performance variation and the subgroup-level patterns, constituted the input for the DIKW learning process.

\begin{table}[t]
\renewcommand{\arraystretch}{0.82}

\centering
\caption{LLM-Predicted vs Actual Message Rankings}
\label{tab:llm-ranks-full}
\footnotesize
\setlength{\tabcolsep}{3pt}
\begin{threeparttable}
\begin{adjustbox}{max width=\textwidth}
\begin{tabular}{@{}lrr rr rr l@{}}
\toprule
& \mc{Actual} & \mc{Actual} & \multicolumn{2}{c}{GPT-4o} & \multicolumn{2}{c}{Claude 3.5} & \\
\cmidrule(lr){4-5} \cmidrule(lr){6-7}
Message & \mc{Rank} & \mc{CTR} & \mc{Rank} & \mc{Err} & \mc{Rank} & \mc{Err} & \mc{Note} \\
\midrule
\textbf{efficiencyTech} & \textbf{1} & \textbf{69.8\%} & \textbf{15} & \textbf{$+14$} & \textbf{14} & \textbf{$+13$} & $\dagger$ \\
salience & 2 & 66.7\% & 6 & $+4$ & 5 & $+3$ &  \\
completePro & 3 & 66.5\% & 3 & 0 & 7 & $+4$ &  \\
\textbf{authorityTrad} & \textbf{4} & \textbf{65.5\%} & \textbf{11} & \textbf{$+7$} & \textbf{19} & \textbf{$+15$} & $\dagger$ \\
progressFeedback & 5 & 65.5\% & 7 & $+2$ & 12 & $+7$ &  \\
\addlinespace
clarityAction & 6 & 65.1\% & 1 & $-5$ & 4 & $-2$ &  \\
\textbf{healthcareStandard} & \textbf{7} & \textbf{64.8\%} & \textbf{13} & \textbf{$+6$} & \textbf{15} & \textbf{$+8$} & $\dagger$ \\
authorityBalance & 8 & 64.2\% & 10 & $+2$ & 13 & $+5$ &  \\
\textbf{avoidSocial} & \textbf{9} & \textbf{64.1\%} & \textbf{16} & \textbf{$+7$} & \textbf{18} & \textbf{$+9$} & $\dagger$ \\
default & 10 & 63.3\% & 14 & $+4$ & 20 & $+10$ &  \\
\addlinespace
tripleTrigger & 11 & 62.0\% & 12 & $+1$ & 1 & $-10$ &  \\
processComplete & 12 & 61.6\% & 4 & $-8$ & 8 & $-4$ &  \\
\textbf{actionDirect} & \textbf{13} & \textbf{61.4\%} & \textbf{2} & \textbf{$-11$} & \textbf{2} & \textbf{$-11$} & $\dagger$ \\
microMessage & 14 & 60.8\% & 20 & $+6$ & 17 & $+3$ &  \\
\textbf{gentleUrgent} & \textbf{15} & \textbf{60.8\%} & \textbf{5} & \textbf{$-10$} & \textbf{3} & \textbf{$-12$} & $\dagger$ \\
\addlinespace
authorityPro & 16 & 59.9\% & 19 & $+3$ & 11 & $-5$ &  \\
cognitiveUltra & 17 & 59.6\% & 18 & $+1$ & 9 & $-8$ &  \\
personalMed & 18 & 57.6\% & 17 & $-1$ & 10 & $-8$ &  \\
personalizationPlus & 19 & 56.9\% & 8 & $-11$ & 16 & $-3$ &  \\
\textbf{reciprocityCue} & \textbf{20} & \textbf{55.5\%} & \textbf{9} & \textbf{$-11$} & \textbf{6} & \textbf{$-14$} & $\dagger$ \\
\midrule
$\rho$ (Spearman) & & & \multicolumn{2}{c}{0.271 (n.s.)} & \multicolumn{2}{c}{-0.120 (n.s.)} & \\
MAE (ranks) & & & \multicolumn{2}{c}{5.7} & \multicolumn{2}{c}{7.7} & Random: 6.7 \\
\bottomrule
\end{tabular}
\end{adjustbox}
\begin{minipage}{\textwidth}
\vspace{0.3em}
\scriptsize
$\dagger$ Consensus failure: both models err by $>$5 ranks (7/20 messages). Predicted ranks from Elo ratings of 380 pairwise comparisons per model. Err = predicted $-$ actual rank.
\end{minipage}
\end{threeparttable}
\end{table}

%
\noindent\textbf{Comparison with Frontier LLMs}
We next examined if frontier LLMs without access to experimental data or analytical tools can predict which interventions would be effective.
Table~\ref{tab:llm-ranks-full} presents the comparison: columns 1--2 show actual Stage~2 performance (discussed in greater detail in the next section) and columns 3--6 show LLM-predicted rankings.
%
We evaluated the performance of two frontier LLMs (GPT-4o \citep{openai2024gpt4o} and Claude 3.5 Sonnet \citep{anthropic2024claude3_5}) without access to Stage~1 experimental data or analytical tools.
%
Each LLM was prompted to assume the persona of a behavioral science expert tasked with predicting patient engagement (prompt in Appendix~\ref{app:llm-prompts}) and instructed to perform pairwise comparisons of all 20 Stage~2 messages (380 comparisons).
%
Results were aggregated using Elo ratings.

%
Neither model achieved meaningful accuracy.
%
Table~\ref{tab:llm-ranks-full} presents the complete comparison across all 20 messages.
%
The best message as shown by the experimental results (efficiencyTech, 69.8\% CTR) was ranked in the bottom third (\#15 GPT-4o, \#14 Claude).
%
The message predicted by both models as \#2 (actionDirect) ranked \#13 in the experiment.
%
Seven consensus failures, where both models erred by more than 5 ranks, reveal systematic miscalibration rather than random noise.
%
Correlation analysis confirms that both models performed poorly: GPT-4o achieved $\rho = 0.271$ ($p = 0.248$, not significant) and Claude $\rho = -0.120$ ($p = 0.613$, not significant).
%
Neither prediction correlated with actual effectiveness.
%
Mean absolute error was 5.70 ranks (GPT-4o) and 7.70 ranks (Claude) versus a 6.7 random baseline; Claude performed worse than random guessing.

%
Based on these findings, we conclude that without access to domain-specific data, LLMs defaulted to cross-domain heuristics from their training data.
%
They over-valued direct commands (actionDirect: predicted rank 2 by both models, actual rank 13) and gentle urgency framing (gentleUrgent: predicted rank 5 by GPT-4o and rank 3 by Claude, actual rank 15), while under-valuing efficiency brevity (efficiencyTech: predicted ranks 15 and 14, actual rank 1) and professional authority (authorityTrad: predicted ranks 11 and 19, actual rank 4), confirming the observation that domain-specific experimental data cannot be substituted by general reasoning \citep{shool2025systematic,liu2024llms}.
%
%
%
%
The poor performance of frontier models in our study parallels a finding from the megastudy literature: when \citet{milkman2024megastudy} asked domain experts to predict which behavioral interventions would be most effective, the experts performed poorly. 
If human experts with deep domain knowledge cannot reliably predict intervention effectiveness \textit{ex ante}, it is perhaps unsurprising that LLMs, which encode a general distillation of published expertise but lack access to the specific experimental context, also fail at this task.

%
These findings establish that the extraction task requires \textit{execution} on data, not \textit{generation} about data: systematically processing 444,691 observations across treatments, subgroups, and interaction effects demands analytical tools, not general reasoning.
%
Next we show how tool-augmented AI with access to the same experimental data produced the best-performing intervention in Stage~2.
%

\subsection{Stage~2 Results}

\begin{table}[t]

\centering
\caption{DIKW-Generated Message Portfolio for Stage~2 Testing}
\label{tab:wisdom-messages}
\scriptsize
\begin{tabular}{p{2.8cm}p{10.2cm}}
\toprule
\textbf{Strategy} & \textbf{Message Variants} \\
\midrule
\textbf{Agentic AI Design} & cognitiveUltra, \textcolor{red}{autonomyMax}, authorityPro, completePro, efficiencyTech, avoidSocial, authorityTrad, tripleTrigger, microMessage, processComplete, personalMed, authorityBalance, actionDirect, gentleUrgent, healthcareStandard, reciprocityCue, \textcolor{red}{microCommitment}, clarityAction, personalizationPlus, \textcolor{red}{stepCompletionUrgency} \\
\textbf{Stage~1 Baselines} & salience, progressFeedback, default \\
\bottomrule
\end{tabular}
\begin{minipage}{\textwidth}
\vspace{0.05cm}
\scriptsize
\textit{Note:} DIKW generated 20 new variants. Red messages (\textcolor{red}{autonomyMax}, \textcolor{red}{microCommitment}, \textcolor{red}{stepCompletionUrgency}) excluded during review; 17 DIKW + 3 baselines = 20 tested. Full message text for every variant is provided in Appendix~\ref{app:stage2-messages}.
\end{minipage}

\end{table}

\begin{figure}[t]
\centering
\begin{subfigure}[t]{0.48\textwidth}
\centering
\includegraphics[width=\textwidth]{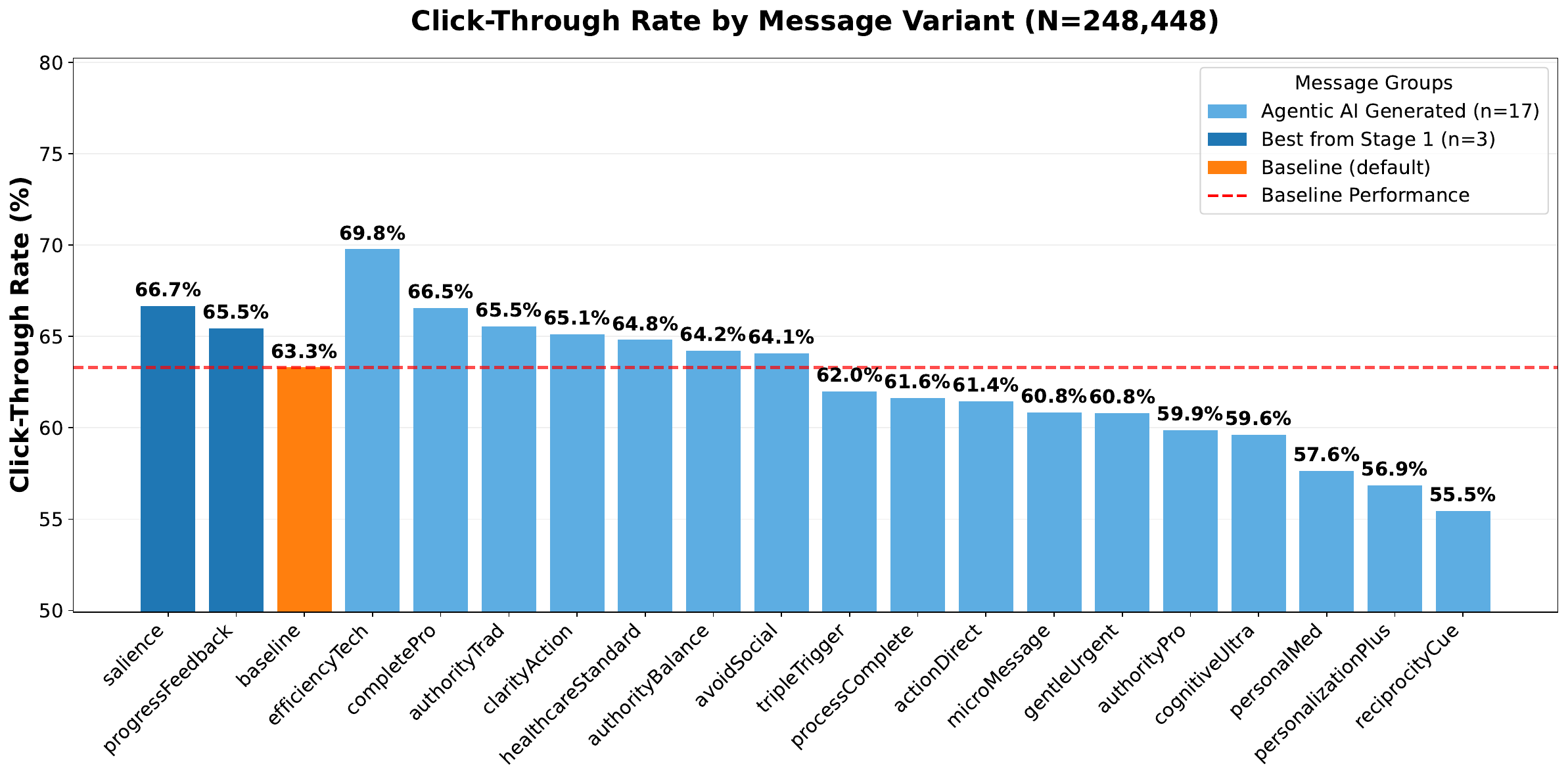}
\caption{Click-Through Rate}
\label{fig:performance-results}
\end{subfigure}
\hfill
\begin{subfigure}[t]{0.48\textwidth}
\centering
\includegraphics[width=\textwidth]{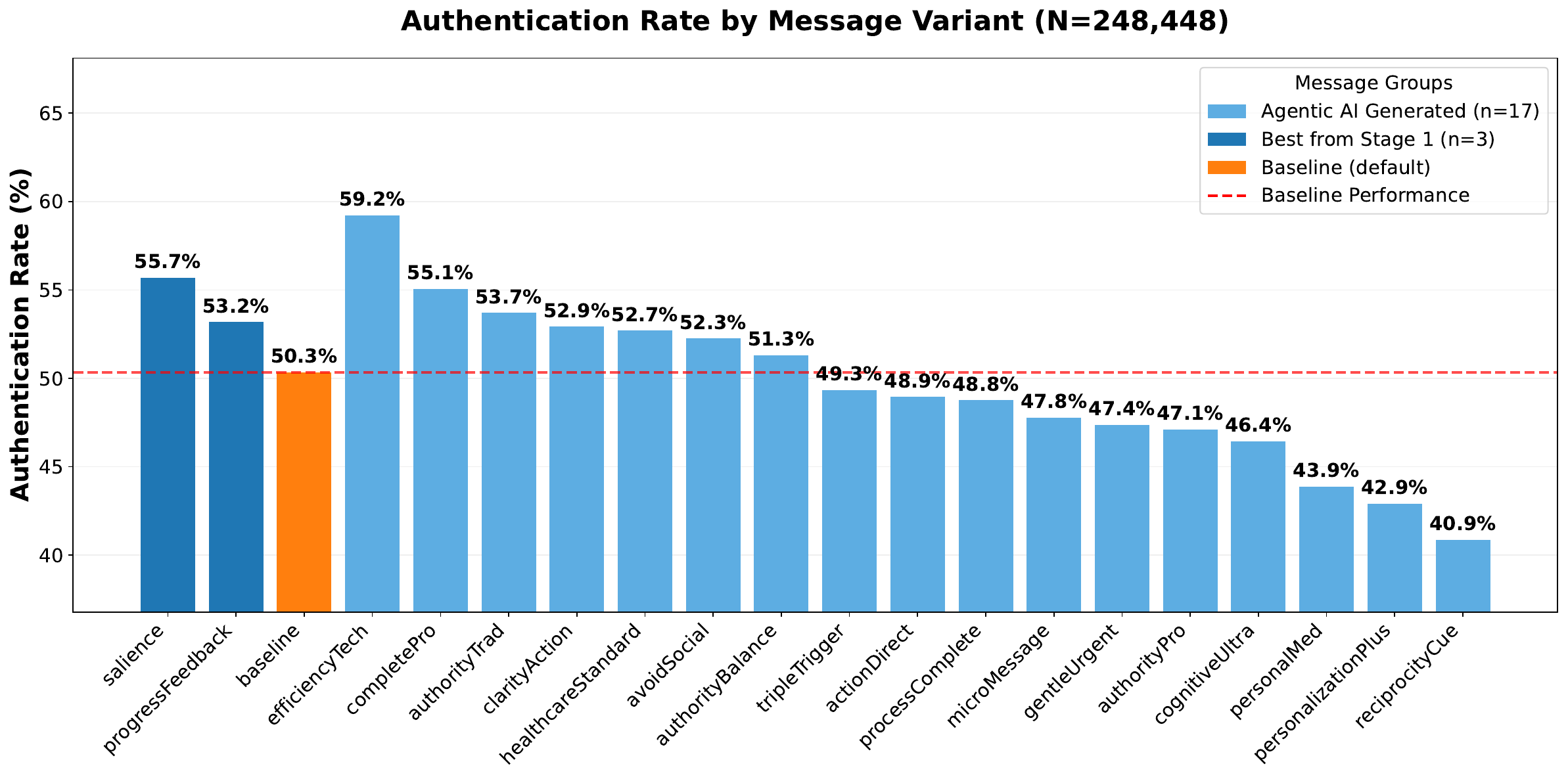}
\caption{Authentication Rate}
\label{fig:s2-auth}
\end{subfigure}
\caption{Stage~2 Message Performance by Variant (N=248,448). (a) Click-through rates across 20 variants grouped by generation strategy. (b) Authentication completion rates confirm engagement patterns persist beyond initial click.}
\label{fig:s2-performance}
\end{figure}

%
To examine whether systematic learning of design knowledge from experimental data can produce superior next-round interventions, we compared the 17 AI-generated messages with the three best performing messages from Stage~1 across 248,448 patient visits. 
%
The answer is yes.
%
Table~\ref{tab:wisdom-messages} lists the 20 message variants tested in Stage~2 (17 AI-generated + 3 Stage~1 baselines, with full message text in Appendix~\ref{app:stage2-messages}) across 248,448 patient visits. Figure~\ref{fig:s2-performance} shows click-through rates for each variant; full regression specifications are in Appendix~\ref{app:regression}.
%
The best-performing message, efficiencyTech, was generated through the DIKW process and achieved 69.8\% CTR, a 6.5 percentage-point improvement over the concurrent baseline (the same default message from Stage~1, retested in Stage~2 at 63.3\% CTR).

\begin{table}[t]

\centering
\caption{Main Treatment Effects on Click-Through Rate}
\label{tab:main-effects-summary}
\scriptsize
\setlength{\tabcolsep}{5pt}
\begin{threeparttable}
\begin{tabular}{l*{7}{c}}
\toprule
\textbf{Message Variant} & \textbf{N} & \mc{\shortstack[t]{Click-\\Through\\Rate (\%)}} & \mc{\shortstack[t]{Authenti-\\cation\\Rate (\%)}} & \mc{\shortstack[t]{(1)\\Simple}} & \mc{\shortstack[t]{(2)\\+Demo}} & \mc{\shortstack[t]{(3)\\+Medical}} & \mc{\shortstack[t]{(4)\\+Temporal}} \\
\midrule
\multicolumn{8}{l}{\textit{\textbf{Agentic AI Generated Messages (17 messages, N=211,173)}}} \\
\quad efficiencyTech         & 12,231 & 69.8 & 59.2 & 0.0654\sym{***} & 0.0645\sym{***} & 0.0645\sym{***} & 0.0644\sym{***} \\
                         &        & & & \mc{(0.0058)} & \mc{(0.0058)} & \mc{(0.0058)} & \mc{(0.0058)} \\
\quad completePro            & 12,305 & 66.5 & 55.1 & 0.0326\sym{***} & 0.0327\sym{***} & 0.0327\sym{***} & 0.0326\sym{***} \\
                         &        & & & \mc{(0.0060)} & \mc{(0.0060)} & \mc{(0.0060)} & \mc{(0.0060)} \\
\quad authorityTrad          & 12,355 & 65.5 & 53.7 & 0.0225\sym{***} & 0.0217\sym{***} & 0.0217\sym{***} & 0.0217\sym{***} \\
                         &        & & & \mc{(0.0060)} & \mc{(0.0060)} & \mc{(0.0060)} & \mc{(0.0060)} \\
\quad clarityAction          & 12,505 & 65.1 & 52.9 & 0.0183\sym{***} & 0.0179\sym{***} & 0.0179\sym{***} & 0.0178\sym{***} \\
                         &        & & & \mc{(0.0060)} & \mc{(0.0060)} & \mc{(0.0060)} & \mc{(0.0060)} \\
\quad \textit{[13 more messages\ldots]} & & & & & & & \\
\\
\multicolumn{8}{l}{\textit{\textbf{Best From Stage 1 (3 messages, N=37,275)}}} \\
\quad salience               & 12,431 & 66.7 & 55.7 & 0.0337\sym{***} & 0.0340\sym{***} & 0.0340\sym{***} & 0.0339\sym{***} \\
                         &        & & & \mc{(0.0060)} & \mc{(0.0059)} & \mc{(0.0059)} & \mc{(0.0059)} \\
\quad progressFeedback       & 12,459 & 65.5 & 53.2 & 0.0216\sym{***} & 0.0219\sym{***} & 0.0219\sym{***} & 0.0219\sym{***} \\
                         &        & & & \mc{(0.0060)} & \mc{(0.0060)} & \mc{(0.0060)} & \mc{(0.0060)} \\
\quad default (reference)   & 12,385 & 63.3 & 50.3 & \mc{--} & \mc{--} & \mc{--} & \mc{--} \\
                         &        & & & \mc{--} & \mc{--} & \mc{--} & \mc{--} \\
\midrule
Demographics            &        & & & \mc{No}  & \mc{Yes} & \mc{Yes} & \mc{Yes} \\
Medical Context         &        & & & \mc{No}  & \mc{No}  & \mc{Yes} & \mc{Yes} \\
Temporal Controls       &        & & & \mc{No}  & \mc{No}  & \mc{No}  & \mc{Yes} \\
\midrule
Observations            &        & & & \mc{248,448} & \mc{248,448} & \mc{248,448} & \mc{248,448} \\
Pseudo R\sym{2}         &        & & & 0.0040 & 0.0127 & 0.0127 & 0.0129 \\
Baseline CTR            &        & & & 0.6330 & 0.6330 & 0.6330 & 0.6330 \\
\bottomrule
\end{tabular}
\begin{minipage}{\linewidth}
\vspace{0.05cm}
\scriptsize
\textit{Note:} Top-performing variants only; full 20-variant breakdown in Appendix~\ref{app:regression} (Table~\ref{tab:main-effects}). Click-Through Rate and Authentication Rate columns report per-arm raw rates. Columns (1)--(4): coefficients from logistic regression of click on variant dummies (baseline = \texttt{default}), with robust standard errors in parentheses. FDR-adjusted (Benjamini--Hochberg) $p$-values applied to column (4). \sym{***} $p<0.01$, \sym{**} $p<0.05$, \sym{*} $p<0.10$.
\end{minipage}
\end{threeparttable}

\end{table}

%
\noindent\textbf{Overall Performance Results}
Click-through rates spanned 14.3 percentage points across the 20 variants (55.5\% to 69.8\%), with the best AI-generated message (efficiencyTech) lifting CTR by 6.5 percentage points (a 10.3\% relative improvement) over the default baseline (63.3\%).
%
This spread is unusually large for an SMS-based health-intervention experiment: comparable megastudies have reported best-arm-versus-control gaps of roughly 2--5 percentage points on binary outcomes, with most candidate treatments not significantly different from baseline \citep{milkman2021megastudy,milkman2022680}.
%
The gain is also not a single-variant fluke: four AI-generated messages (efficiencyTech, completePro, authorityTrad, and clarityAction) significantly outperformed the default baseline after Holm-Bonferroni correction (all $p \leq 0.0031$), indicating that the DIKW pipeline produces a portfolio of effective interventions rather than a lucky draw.
%
Most importantly, the best AI-generated message also outperformed the strongest message from the Stage~1 human-plus-LLM portfolio (salience, 66.7\%) by 3.1 percentage points (4.6\% relative).
%
The agentic DIKW process therefore exceeds the current state of practice in AI-assisted message design, not merely the existing standard of care.
%
Authentication completion rates show the same ordering (Figure~\ref{fig:s2-performance}(b)), confirming that the engagement gains persist into the deeper, identity-verified step of the funnel rather than dissipating at the click.

%
We estimate the click-through model
\begin{equation}
\Pr(\text{click}_i = 1 \mid v_i, X_i) = \Lambda(\alpha + \beta_{v_i} + X_i'\gamma),
\label{eq:s2-logit}
\end{equation}
where $\Lambda(\cdot)$ is the logistic CDF, $v_i$ indexes the message variant assigned to invitation $i$, and $\beta_{\text{default}} = 0$ is the omitted reference.
%
The covariate vector $X_i$ varies across the four specifications shown in Table~\ref{tab:main-effects-summary}: column~(1) has $X_i = \emptyset$, and columns~(2)--(4) progressively add demographic, medical, and temporal controls.
%
Reported coefficients are estimated relative to \texttt{default}, with robust standard errors.
%
Table~\ref{tab:main-effects-summary} reports per-variant treatment effects on click-through rate from a logistic regression with demographic, medical, and temporal controls.
%
The effects align with the raw rate differences shown in Figure~\ref{fig:s2-performance}.
%
Coefficients are stable across progressively richer specifications, with the full four-column robustness check reported in Appendix~\ref{app:regression}.
%
How did the DIKW system generate these superior interventions? Below we trace the full set of behavioral principles extracted by the system and their connection to these Stage~2 results.
%

\subsection{Transparency of the Agentic AI Learning Process}

\begin{figure}[t]
\centering
\includegraphics[width=\textwidth]{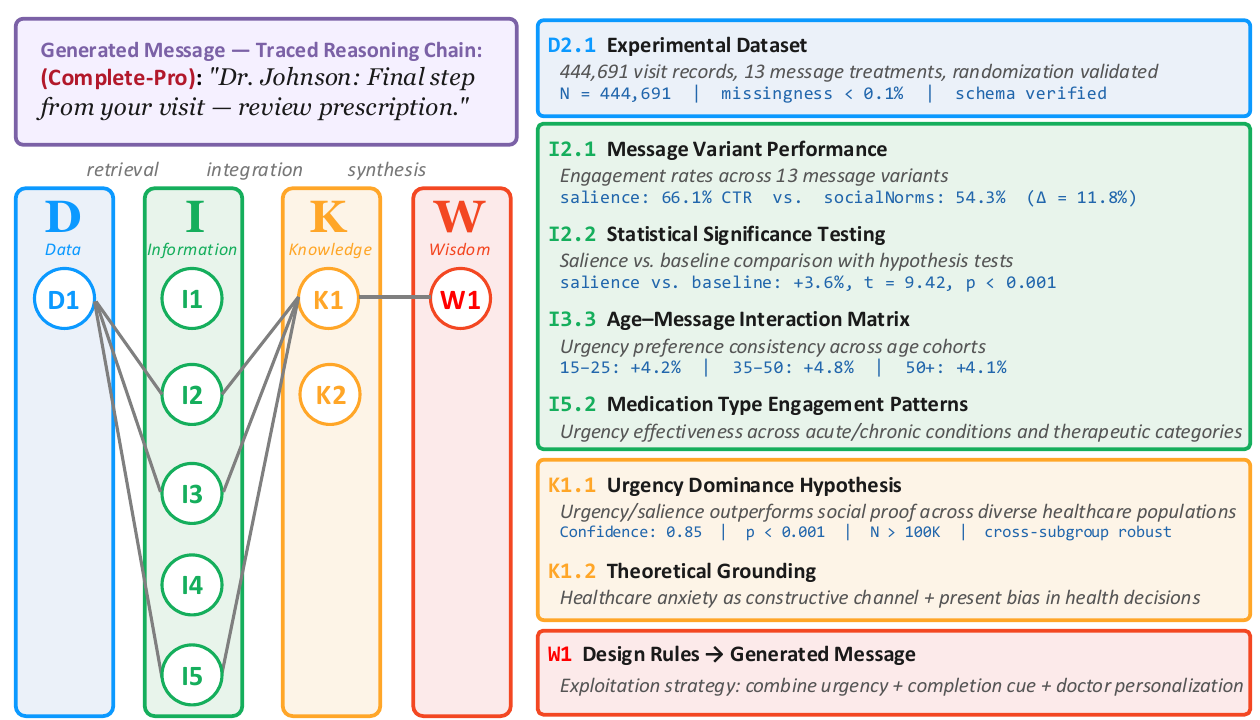}
\caption{DIKW Evidence Chain Trace: From Urgency-Salience Pattern to completePro Message Generation. This trace demonstrates how the system maintains explicit evidence chains across all four DIKW levels, linking the final message design back to specific statistical patterns observed in Stage~1 data.}
\label{fig:evidence-chain}
\end{figure}

%
One major advantage of our agentic AI approach is the transparency and explainability in performance.
%
The learning process began with the orchestrator agent determining relevant analytical topics at each DIKW level (Table~\ref{tab:dikw-topics}).
We trace the subsequent execution through each level, from data validation through statistical pattern discovery to knowledge synthesis and message design (Figure~\ref{fig:evidence-chain}).

\begin{table}[t]

\centering
\caption{DIKW Layer Topics and Processing Results}
\label{tab:dikw-topics}
\scriptsize
\begin{tabular}{p{2cm}p{10.5cm}}
\toprule
\textbf{Layer} & \textbf{Topics Processed} \\
\midrule
\textbf{Data} & Dataset Description, Experiment Description, Data Quality Validation, Missing Value Analysis, Schema Verification \\
\textbf{Information} & Engagement Fundamentals, Message Performance, Patient Demographics, Temporal Dynamics, Medical Context, Geographic Context, Message Strategy × Demographics, Message Strategy × Medical Context, Message Strategy × Temporal Context, Linguistic Features × Context, Message Strategy × Geographic/SES, Cross-Strategy Performance Patterns \\
\textbf{Knowledge} & Psychological Messaging Principles, Patient Segmentation Strategy, Healthcare Communication Timing, Trust and Authority Dynamics, Medical Context Adaptation, Behavioral Economics in Healthcare, Message Strategy Optimization Frameworks, Linguistic Optimization Principles, Sequential Message Strategy, Contextual Sensitivity Patterns, Behavioral Prediction Models, Cross-Cultural Healthcare Communication \\
\bottomrule
\end{tabular}

\end{table}

The D-Agent began at the Data level, where it built an understanding of the experimental context through autonomous code execution.
%
It documented the dataset structure, identifying that each of the 444,691 records represented a single patient prescription visit, and cataloged the available features by category: patient demographics (age, gender, location), medical context (drug category, provider specialty), temporal patterns (time of day, day of week), and engagement outcomes (click, authentication).
%
The D-Agent also inventoried the 13 message variants along with their specific text, and verified data quality, confirming balanced randomization across treatment arms.
%
This foundational understanding of the data, the experimental design, and the intervention content informed all subsequent analysis.

%
Building on this understanding, the I-Agent's Information-level analyses produced a set of individual statistical findings from the Stage~1 data, covering engagement metrics, demographic patterns, medical context, and temporal dynamics.
%
As presented earlier, click-through rates varied substantially across the 13 message variants, ranging from 54.3\% (socialNorms) to 66.1\% (salience) against a default baseline of 62.5\%.
%
Subgroup analyses revealed that message length correlated negatively with engagement; that the top-performing messages shared action-oriented language (``review,'' ``new''); that age moderated effectiveness, with younger patients showing larger treatment effects; and that social proof messaging (``most patients find this useful'') underperformed the baseline among both male and female patients, across all age groups, and in every medical context the agent examined.
%
Each of these was an individual observation about the Stage~1 data, produced through autonomous code execution and preserved as a reproducible statistical report.

%
At the Knowledge level, the K-Agent synthesized these individual findings into integrated assessments of which messaging principles are reliable in this domain.
%
No single Information-level finding established this on its own; rather, the K-Agent combined converging evidence across multiple analyses.
%
For instance, the K-Agent identified efficiency framing as a reliable principle by integrating several findings: the negative correlation between message length and engagement, the higher performance of action-oriented language, and the consistency of these patterns across age, gender, and medical-context subgroups.
%
Similarly, the K-Agent concluded that social proof is ineffective in this healthcare context by synthesizing its consistently lower performance across every subgroup analysis, leading it to exclude social proof from message generation entirely.
%
The K-Agent also identified completion framing and professional authority as consistent patterns supported by multiple Information-level analyses, though with smaller effect sizes.
%
Each Knowledge-level assessment was thus grounded in the convergence of multiple Information-level findings, maintaining explicit evidence chains from design decisions back to specific statistical observations (Figure~\ref{fig:evidence-chain}; a complete interpretability trace is provided in Appendix~\ref{app:agent-outputs}).

%
At the Wisdom level, the W-Agent translated its Knowledge-level assessments into concrete message designs that satisfied the same operational constraints as Stage~1 (maximum 84 characters, provider attribution, no misleading claims), which were then tested in Stage~2 with 248,448 new patient visits.
%
Based on the efficiency-framing assessment, the W-Agent designed efficiencyTech (``New Rx info needs quick review''), which became the best-performing message.
%
Based on completion framing, the W-Agent designed completePro (``Final step from your visit; review prescription''), which achieved 66.5\% CTR (+3.2 percentage points).
%
Based on professional authority, the W-Agent designed authorityTrad (``Dr. [Provider Name] requests: Please review your prescription''), which achieved 65.5\% CTR (+2.2 percentage points).
%
The W-Agent also designed messages that tested mechanisms beyond what Stage~1 data could directly support: reciprocityCue tested whether expressing gratitude before action would engage patients, and personalizationPlus adopted a casual first-name tone.
%
Both performed poorly (reciprocityCue: $-$7.8 percentage points; personalizationPlus: $-$6.4 percentage points), suggesting that the agent's designs were most effective when grounded in reliable Stage~1 evidence.

%
To unpack why efficiencyTech leads, we note that two ingredients distinguish it from lower-ranked variants.
%
First, brevity (54 characters, well below the 84-character limit) aligns with the I-level finding that message length correlates negatively with engagement.
%
Second, the pairing of a novelty cue (``New'') with a concrete action verb (``review'') mirrors the action-oriented language pattern the I-Agent identified among Stage~1 top performers.
%
We also found that variation in message performance across age (efficiencyTech: $+9.1$ percentage points for 18--44 vs.\ $+4.0$ percentage points for 45--64; interaction $p<0.001$) is consistent with cognitive-load reduction mattering more for digitally-native patients.

%
The agent's learning process produced domain-specific knowledge by synthesizing fragmented statistical findings into integrated assessments of which messaging principles work in this healthcare context.
%
Efficiency framing, completion framing, and professional authority emerged as reliable principles; social proof, one of the most widely applied nudging techniques \citep{cialdini2007descriptive}, was identified as ineffective and excluded.
%
The agent's output was not merely a ranking of which messages performed best, but a structured assessment of which principles are reliable in this domain, translated into concrete intervention designs.
%
We develop the theoretical interpretation of why these domain-specific patterns emerge in Section~5.3.
%
To summarize, the agentic AI workflow advances transparency and auditability through three architectural choices: separating objective code execution from LLM reasoning, requiring multi-finding convergence at the Knowledge level, and maintaining explicit evidence chains from each design decision back to Stage~1 observations.
%


\section{Discussion}

%
Our study demonstrates that behavioral experiments can function as cumulative knowledge-building systems when organizations invest in learning infrastructure that combines domain-specific data, analytical tools, and structured reasoning.
%
Two-stage field experiments with 693,139 patient visits validated this: the DIKW process produced the best-performing intervention and identified which behavioral principles apply in this healthcare context and which do not, resolving uncertainty that general theory alone leaves open.

%
We make four contributions to the domain of learning from behavioral experiments in healthcare messaging.
%
\textbf{Cumulative Learning:} To the best of our knowledge, this is the first study to show that tool-augmented agentic AI can learn from experimental data to generate superior interventions, demonstrating that experiments can function as cumulative knowledge systems.
%
\textbf{Computational Grounding:} We are able to show that the learning advantage requires systematic code execution on domain-specific experimental data, not AI reasoning alone. 
%
The DIKW architecture grounds LLM synthesis in verified computational outputs: by separating code execution (D/I-levels, producing reproducible statistical evidence) from knowledge synthesis (K/W-levels, reasoning over that evidence), the system prevents hallucination of behavioral patterns that unaided LLMs generate from general training data.
%
\textbf{Design Science:} The DIKW multi-agent architecture provides validated design principles for building learning systems, including transparency and explainability through evidence chains linking every design decision back to specific experimental observations.
%
\textbf{Knowledge Discovery:} The system identifies which behavioral principles apply in a given domain, resolving the theory selection problem that general nudging frameworks leave open.
%
We discuss these contributions in more detail below.

\noindent\textbf{Cumulative Learning and the Role of Infrastructure}

%
The central finding is that tool-augmented agentic AI, equipped with domain-specific experimental data and structured DIKW reasoning (Section~2.3), produces measurably superior interventions from the same data that unaided reasoning cannot meaningfully interpret.
%
The DIKW hierarchy explains why: each level of abstraction (data validation, pattern extraction, knowledge synthesis, design generation) requires distinct computational capabilities, and spanning all four levels with integrated tools produces knowledge that no single approach achieves alone.

%
A natural concern is whether Stage~2's improvement simply reflects having more information, or if human experts could have extracted similar knowledge from the same data.
%
We make three observations that address both concerns.
%
%
First, organizations systematically fail to extract knowledge from experiments despite having the data, because scalable learning infrastructure has been absent \citep{levitt1988organizational,argote2012organizational}.
%
Second, the DIKW architecture makes extraction feasible at megastudy scale, processing a combinatorial space of treatments, subgroups, and interactions that exceeds human cognitive capacity.
%
Third, the extracted knowledge is substantively non-obvious: general behavioral theory would not have suggested that efficiency framing dominates or that social proof backfires in healthcare.
%
Consistent with \citet{milkman2024megastudy}'s finding that domain experts asked to predict intervention effectiveness also performed poorly; neither human expertise nor frontier AI reasoning can substitute for systematic extraction from domain-specific experimental data. 
%
This extends prior work on AI in organizations \citep{agrawal2022prediction}: AI's value extends beyond prediction to systematic knowledge creation, but only when equipped with domain-specific data and analytical infrastructure.
%

%
A second concern is whether the Stage~2 effects themselves are contaminated by repeat messaging or prior exposure within the experiment window.
In practice, repeat exposure is rare.
Only 1.29\% of Stage~2 patients (3{,}164 of 245{,}227) received more than one invitation, and 84.95\% had no SMS messages in the prior 7~days.
Because randomization is independent at the invitation level, any contamination would have to operate through within-patient carryover, and the data show no such effect.
Multi-visit patients have slightly \emph{higher} CTR than their single-visit counterparts (64.70\% vs.\ 62.52\%), and previously-exposed patients have slightly higher CTR than unexposed patients (64.93\% vs.\ 62.16\%).

\noindent\textbf{Architecture and Design Principles}

%
The design science contribution is a set of architectural principles for building learning systems, validated through the DIKW implementation and Stage~2 field results \citep{hevner2004design,gregor2013positioning}.
%
\textit{Hierarchical abstraction} prevents the system from jumping directly from raw data to intervention designs, requiring intermediate steps of pattern extraction and knowledge synthesis.
%
\textit{Transparency and explainability} are achieved through evidence chain preservation: every assessment traces back through specific statistical findings to raw observations, creating auditable reasoning.
%
\textit{Bidirectional flow} combines bottom-up discovery (patterns emerging from data) with top-down refinement (targeted analyses triggered by insufficient evidence), distinguishing this from simple data mining.
%
\textit{Computational grounding} separates objective code execution (D/I-levels, producing reproducible statistical facts) from subjective synthesis (K/W-levels, LLM reasoning over those facts). 
This prevents hallucination: the synthesizing agent cannot assert a behavioral pattern that the code execution component has not already verified numerically. 
This principle distinguishes tool-augmented agentic AI from approaches that simply provide an LLM with a data summary and rely on its reasoning alone.
%
While validated in healthcare, these design principles are domain-independent: any setting where organizations accumulate experimental data but fail to extract cumulative knowledge presents the same architectural challenge.

\noindent\textbf{Why Domain-Specific Knowledge Emerges}

%
The strongest evidence that the learning method works is the nature of the knowledge it produces: domain-specific principles that general behavioral theory does not predict.
%
Drawing on the Stage~2 results (Section~4.3) and prior healthcare communication research, we identify three domain-specific factors that explain why widely used behavioral principles backfire in this context.

%
First, \textit{autonomy preservation}: patients resist directive language in medical communications, consistent with reactance theory \citep{brehm1966theory}; this explains why both direct commands and explicit urgency framing underperformed, while efficiency framing conveyed urgency implicitly without triggering resistance.
%
Second, \textit{professional norm governance}: a casual tone undermines credibility in medical contexts where competence signaling matters \citep{pornpitakpan2004persuasiveness}; this explains why professional authority succeeds while casual personalization backfires.
%
Third, \textit{cognitive load sensitivity}: patients facing competing medical demands reward brevity and clarity over persuasive elaboration; this explains why efficiency framing produces the strongest effects, and why completion framing succeeds by positioning the action as finishing an existing process rather than starting a new task \citep{zeigarnik1938finished}.
%
Together, these factors also explain the failure of social proof (patients resist conformity pressure about personal health decisions \citep{cialdini2007descriptive}) and reciprocity (premature gratitude triggers the persuasion knowledge model \citep{friestad1994persuasion}).

%
These findings demonstrate that domain-specific knowledge creation requires domain-specific experimental data processed through learning infrastructure; cross-domain theory transfer is insufficient.

\noindent\textbf{Implications, Limitations, and Future Research}

\noindent\textbf{Practical Implications}
%
%
For organizations conducting behavioral experiments, the primary implication is to invest in learning infrastructure rather than treating each experiment as an isolated evaluation.
%
This means integrating systematic learning into experimental workflows: every experiment should feed the next round's design through structured knowledge extraction, not simply identify winners for deployment.
%
For behavioral science, the results suggest a shift from one-shot hypothesis testing toward iterative learning systems where AI serves as a systematic learning partner, building domain-specific knowledge libraries across successive experimental rounds.

\noindent\textbf{Domain and Context Generalizability}
%
%
The systematic learning approach was validated in healthcare prescription notifications with high trust requirements and autonomy sensitivity.
%
The question of whether it generalizes to domains with different dynamics (consumer marketing, financial services, education) remains open.
%
Related work in mobile contexts shows that targeting and design effects vary substantially with situational and trajectory factors \citep{andrews2016mobile,ghose2019mobile}, suggesting that principles extracted in our healthcare setting may not transfer directly even within digital messaging domains.
%
We hypothesize the architecture generalizes while behavioral principles vary by domain; future research should test this across contexts.

\noindent\textbf{Outcome Measurement Scope}
%
%
Across both stages, the DIKW principles produce consistent rankings on the two engagement outcomes we observe: click-through (the primary measure) and authentication (the secondary, deeper measure of identity verification after the click).
%
This consistency indicates that the extracted design principles reach beyond initial attention and propagate to a more committed downstream step.
%
Future research should extend the funnel further, to outcomes such as prescription pickup, refill behavior, and clinical adherence.
%
Our evaluation focuses on short-term engagement (click-through within 24 hours) rather than downstream health outcomes (medication adherence, clinical endpoints).
%
Future research should extend to sustained behavior change and assess potential negative effects (patient stress, decision fatigue, trust erosion).
%
Privacy and consent considerations also remain salient in healthcare data infrastructures \citep{adjerid2016impact} and in field-experiment contexts more broadly \citep{godinho2022consumer}.

\noindent\textbf{Human Expert Comparison}
%
%
It may be possible that, with sufficient resources and availability of expertise, a similar learning process could be replicated by human analysts given full access to Stage~1 statistical outputs. 
A controlled comparison with expert analysts remains an important direction for future research.
%
However, this study demonstrates the value of an automated pipeline that achieves significant efficiency gains: the DIKW system processed the full combinatorial space of treatments, subgroups, and interactions in a matter of hours, a task that would require substantial time and expertise from a human team. 
This efficiency makes systematic experimental learning accessible to organizations that lack dedicated research staff, contributing to the democratization of evidence-based intervention design.

\noindent\textbf{Knowledge Validation}
%
%
The current study does not formally test whether the agent's confidence assessments predict Stage~2 outcomes at the individual-hypothesis level.
%
Future research could design experiments that directly test this relationship, providing stronger evidence for the value of structured confidence scoring in AI-assisted learning systems.

\section{Conclusion}

%
This study demonstrates that behavioral experiments can function as cumulative knowledge-building systems when organizations equip AI with domain-specific experimental data, analytical tools, and structured reasoning.
%
%
The path to AI value in organizational learning runs not through more capable models but through learning infrastructure that connects AI to domain-specific experimental data.
%
Emerging developments in AI, particularly tool-augmented agentic systems with code execution and structured reasoning, open new possibilities for both scientific discovery and pragmatic intervention design.
%
As organizations accumulate experimental evidence across domains, the bottleneck shifts from data collection to knowledge extraction, and the DIKW architecture offers a replicable template for closing that gap.

\begingroup
  \parindent 0pt \parskip 0.0ex
  \def\enotesize{\normalsize}
  \theendnotes
\endgroup

\bibliographystyle{informs2014}
\bibliography{reference}

\begin{thebibliography}{75}
\providecommand{\natexlab}[1]{#1}
\providecommand{\url}[1]{\texttt{#1}}
\providecommand{\urlprefix}{URL }

\bibitem[{Ackoff(1989)}]{ackoff1989data}
Ackoff RL (1989) From data to wisdom. \emph{Journal of applied systems
  analysis} 16(1):3--9.

\bibitem[{Adjerid et~al.(2016)Adjerid, Acquisti, Telang, Padman,
  \protect\BIBand{} Adler-Milstein}]{adjerid2016impact}
Adjerid I, Acquisti A, Telang R, Padman R, Adler-Milstein J (2016) The impact
  of privacy regulation and technology incentives: The case of health
  information exchanges. \emph{Management Science} 62(4):1042--1063.

\bibitem[{Adjerid et~al.(2018)Adjerid, Peer, \protect\BIBand{}
  Acquisti}]{adjerid2018beyond}
Adjerid I, Peer E, Acquisti A (2018) Beyond the privacy paradox: Objective
  versus relative risk in privacy decision making1. \emph{MIS quarterly}
  42(2):465--488.

\bibitem[{Agrawal et~al.(2022)Agrawal, Gans, \protect\BIBand{}
  Goldfarb}]{agrawal2022prediction}
Agrawal A, Gans J, Goldfarb A (2022) \emph{Prediction machines, updated and
  expanded: The simple economics of artificial intelligence} (Harvard Business
  Press).

\bibitem[{Alavi \protect\BIBand{} Leidner(1999)}]{alavi1999knowledge}
Alavi M, Leidner DE (1999) Knowledge management and knowledge management
  systems. \emph{MIS Quarterly} 25.

\bibitem[{Almaatouq et~al.(2024)Almaatouq, Griffiths, Suchow, Whiting, Evans,
  \protect\BIBand{} Watts}]{almaatouq2024beyond}
Almaatouq A, Griffiths TL, Suchow JW, Whiting ME, Evans J, Watts DJ (2024)
  Beyond playing 20 questions with nature: Integrative experiment design in the
  social and behavioral sciences. \emph{Behavioral and Brain Sciences} 47:e33.

\bibitem[{Andrews et~al.(2016)Andrews, Luo, Fang, \protect\BIBand{}
  Ghose}]{andrews2016mobile}
Andrews M, Luo X, Fang Z, Ghose A (2016) Mobile ad effectiveness:
  Hyper-contextual targeting with crowdedness. \emph{Marketing Science}
  35(2):218--233.

\bibitem[{{Anthropic}(2024)}]{anthropic2024claude3_5}
{Anthropic} (2024) Claude 3.5 sonnet.
  \url{https://www.anthropic.com/news/claude-3-5-sonnet}, accessed: October
  2025.

\bibitem[{{Anthropic}(2025)}]{anthropic2025claude4}
{Anthropic} (2025) Introducing claude 4: Opus 4 and sonnet 4.
  \urlprefix\url{https://www.anthropic.com/news/claude-4}, blog announcement
  introducing Claude 4 models (Opus 4 and Sonnet 4).

\bibitem[{Argote(2012)}]{argote2012organizational}
Argote L (2012) \emph{Organizational learning: Creating, retaining and
  transferring knowledge} (Springer Science \& Business Media).

\bibitem[{Bennett \protect\BIBand{} Metz(2024)}]{bloomberg2024openai}
Bennett D, Metz R (2024) {OpenAI} tells staff it has reached level 2 of 5 in
  {AI} development. Bloomberg, reports OpenAI's internal five-level AGI
  framework: Level~1 Chatbots, Level~2 Reasoners, Level~3 Agents, Level~4
  Innovators, Level~5 Organizations.

\bibitem[{Brehm(1966)}]{brehm1966theory}
Brehm JW (1966) A theory of psychological reactance. \emph{Nova lorque.
  Academic Press} .

\bibitem[{Brown \protect\BIBand{} Bussell(2011)}]{brown2011medication}
Brown MT, Bussell JK (2011) Medication adherence: Who cares? \emph{Mayo clinic
  proceedings}, volume~86, 304--314 (Elsevier).

\bibitem[{Brynjolfsson et~al.(2025)Brynjolfsson, Li, \protect\BIBand{}
  Raymond}]{brynjolfsson2025generative}
Brynjolfsson E, Li D, Raymond L (2025) Generative ai at work. \emph{The
  Quarterly Journal of Economics} 140(2):889--942.

\bibitem[{Chang et~al.(2017)Chang, DeVore, Granger, Eapen, Ariely,
  \protect\BIBand{} Hernandez}]{chang2017leveraging}
Chang LL, DeVore AD, Granger BB, Eapen ZJ, Ariely D, Hernandez AF (2017)
  Leveraging behavioral economics to improve heart failure care and outcomes.
  \emph{Circulation} 136(8):765--772.

\bibitem[{Chen \protect\BIBand{} Chan(2024)}]{chen2024large}
Chen Z, Chan J (2024) Large language model in creative work: The role of
  collaboration modality and user expertise. \emph{Management Science}
  70(12):9101--9117.

\bibitem[{Cialdini(2007)}]{cialdini2007descriptive}
Cialdini RB (2007) Descriptive social norms as underappreciated sources of
  social control. \emph{Psychometrika} 72(2):263--268.

\bibitem[{Cialdini et~al.(2009)}]{cialdini2009influence}
Cialdini RB, et~al. (2009) \emph{Influence: Science and practice}, volume~4
  (Pearson education Boston).

\bibitem[{Czaja et~al.(2006)Czaja, Charness, Fisk, Hertzog, Nair, Rogers,
  \protect\BIBand{} Sharit}]{czaja2006factors}
Czaja SJ, Charness N, Fisk AD, Hertzog C, Nair SN, Rogers WA, Sharit J (2006)
  Factors predicting the use of technology: findings from the center for
  research and education on aging and technology enhancement (create).
  \emph{Psychology and aging} 21(2):333.

\bibitem[{Dell'Acqua et~al.(2023)Dell'Acqua, McFowland~III, Mollick,
  Lifshitz-Assaf, Kellogg, Rajendran, Krayer, Candelon, \protect\BIBand{}
  Lakhani}]{dell2023navigating}
Dell'Acqua F, McFowland~III E, Mollick ER, Lifshitz-Assaf H, Kellogg K,
  Rajendran S, Krayer L, Candelon F, Lakhani KR (2023) Navigating the jagged
  technological frontier: Field experimental evidence of the effects of ai on
  knowledge worker productivity and quality. \emph{Harvard business school
  technology \& operations mgt. Unit working paper} (24-013).

\bibitem[{Duckworth et~al.(2022)Duckworth, Milkman, \protect\BIBand{}
  Nelson}]{duckworth2022guide}
Duckworth AL, Milkman KL, Nelson KE (2022) A guide to megastudies. \emph{PNAS
  nexus} 1(5):pgac214.

\bibitem[{Feng et~al.(2022)Feng, Claggett, Karahanna, \protect\BIBand{}
  Tam}]{feng2022randomized}
Feng Y, Claggett JL, Karahanna E, Tam KY (2022) A randomized field experiment
  to explore the impact of herding cues as catalysts for adoption. \emph{Mis
  Quarterly} 46(2):1135--1164.

\bibitem[{Friestad \protect\BIBand{} Wright(1994)}]{friestad1994persuasion}
Friestad M, Wright P (1994) The persuasion knowledge model: How people cope
  with persuasion attempts. \emph{Journal of consumer research} 21(1):1--31.

\bibitem[{Ghose et~al.(2022)Ghose, Guo, Li, \protect\BIBand{}
  Dang}]{ghose2022empowering}
Ghose A, Guo X, Li B, Dang Y (2022) Empowering patients using smart mobile
  health platforms: Evidence from a randomized field experiment. \emph{MIS
  Quarterly} 46(1):151--192.

\bibitem[{Ghose et~al.(2024)Ghose, Lee, Nam, \protect\BIBand{}
  Oh}]{ghose2024effects}
Ghose A, Lee HA, Nam K, Oh W (2024) The effects of pressure and self-assurance
  nudges on product purchases and returns in online retailing: Evidence from a
  randomized field experiment. \emph{Journal of Marketing Research}
  61(3):517--535.

\bibitem[{Ghose et~al.(2019)Ghose, Li, \protect\BIBand{} Liu}]{ghose2019mobile}
Ghose A, Li B, Liu S (2019) Mobile targeting using customer trajectory
  patterns. \emph{Management Science} 65(11):5027--5049.

\bibitem[{Godinho~de Matos \protect\BIBand{}
  Adjerid(2022)}]{godinho2022consumer}
Godinho~de Matos M, Adjerid I (2022) Consumer consent and firm targeting after
  gdpr: The case of a large telecom provider. \emph{Management Science}
  68(5):3330--3378.

\bibitem[{Gregor \protect\BIBand{} Hevner(2013)}]{gregor2013positioning}
Gregor S, Hevner AR (2013) Positioning and presenting design science research
  for maximum impact1. \emph{MIS quarterly} 37(2):337--355.

\bibitem[{Gupta(2011)}]{gupta2011intention}
Gupta SK (2011) Intention-to-treat concept: a review. \emph{Perspectives in
  clinical research} 2(3):109--112.

\bibitem[{Hevner et~al.(2004)Hevner, March, Park, \protect\BIBand{}
  Ram}]{hevner2004design}
Hevner AR, March ST, Park J, Ram S (2004) Design science in information systems
  research1. \emph{MIS quarterly} 28(1):75--106.

\bibitem[{Ho et~al.(2025)Ho, Glorioso, Allen, Blankenhorn, Glasgow, Grunwald,
  Khanna, Magid, Marrs, Novins-Montague et~al.}]{ho2025personalized}
Ho PM, Glorioso TJ, Allen LA, Blankenhorn R, Glasgow RE, Grunwald GK, Khanna A,
  Magid DJ, Marrs J, Novins-Montague S, et~al. (2025) Personalized patient data
  and behavioral nudges to improve adherence to chronic cardiovascular
  medications: a randomized pragmatic trial. \emph{JAMA} 333(1):49--59.

\bibitem[{Hong et~al.(2025)Hong, Lin, Liu, Liu, Wu, Zhang, Li, Chen, Zhang,
  Wang et~al.}]{hong2025data}
Hong S, Lin Y, Liu B, Liu B, Wu B, Zhang C, Li D, Chen J, Zhang J, Wang J,
  et~al. (2025) Data interpreter: An llm agent for data science. \emph{Findings
  of the Association for Computational Linguistics: ACL 2025}, 19796--19821.

\bibitem[{Hou et~al.(2025)Hou, Wang, Wang, Wang, \protect\BIBand{}
  Yang}]{hou2025double}
Hou J, Wang L, Wang G, Wang HJ, Yang S (2025) The double-edged roles of
  generative ai in the creative process: experiments on design work.
  \emph{Information Systems Research} .

\bibitem[{Jennex(2009)}]{jennex2009re}
Jennex ME (2009) Re-visiting the knowledge pyramid. \emph{2009 42nd Hawaii
  International Conference on System Sciences}, 1--7 (IEEE).

\bibitem[{Ji et~al.(2023)Ji, Lee, Frieske, Yu, Su, Xu, Ishii, Bang, Madotto,
  \protect\BIBand{} Fung}]{ji2023survey}
Ji Z, Lee N, Frieske R, Yu T, Su D, Xu Y, Ishii E, Bang YJ, Madotto A, Fung P
  (2023) Survey of hallucination in natural language generation. \emph{ACM
  computing surveys} 55(12):1--38.

\bibitem[{Jimenez et~al.(2023)Jimenez, Yang, Wettig, Yao, Pei, Press,
  \protect\BIBand{} Narasimhan}]{jimenez2023swe}
Jimenez CE, Yang J, Wettig A, Yao S, Pei K, Press O, Narasimhan K (2023)
  Swe-bench: Can language models resolve real-world github issues? \emph{arXiv
  preprint arXiv:2310.06770} .

\bibitem[{Kahneman(2011)}]{kahneman2011thinking}
Kahneman D (2011) \emph{Thinking, Fast and Slow} (New York: Farrar, Straus and
  Giroux).

\bibitem[{Kahneman \protect\BIBand{} Klein(2009)}]{kahneman2009conditions}
Kahneman D, Klein G (2009) Conditions for intuitive expertise: a failure to
  disagree. \emph{American psychologist} 64(6):515.

\bibitem[{Kahneman \protect\BIBand{} Tversky(2013)}]{kahneman2013prospect}
Kahneman D, Tversky A (2013) Prospect theory: An analysis of decision under
  risk. \emph{Handbook of the fundamentals of financial decision making: Part
  I}, 99--127 (World Scientific).

\bibitem[{Kohavi et~al.(2020)Kohavi, Tang, \protect\BIBand{}
  Xu}]{kohavi2020trustworthy}
Kohavi R, Tang D, Xu Y (2020) \emph{Trustworthy online controlled experiments:
  A practical guide to a/b testing} (Cambridge University Press).

\bibitem[{Kwame \protect\BIBand{} Petrucka(2021)}]{kwame2021literature}
Kwame A, Petrucka PM (2021) A literature-based study of patient-centered care
  and communication in nurse-patient interactions: barriers, facilitators, and
  the way forward. \emph{BMC nursing} 20(1):158.

\bibitem[{{LangChain AI}(2025)}]{langgraph2025}
{LangChain AI} (2025) {LangGraph}: Low-level orchestration framework for
  building stateful agents. Computer software,
  \urlprefix\url{https://github.com/langchain-ai/langgraph}.

\bibitem[{Levitt \protect\BIBand{} March(1988)}]{levitt1988organizational}
Levitt B, March JG (1988) Organizational learning. \emph{Annual review of
  sociology} 14(1):319--338.

\bibitem[{Lewis et~al.(2020)Lewis, Perez, Piktus, Petroni, Karpukhin, Goyal,
  K{\"u}ttler, Lewis, Yih, Rockt{\"a}schel et~al.}]{lewis2020retrieval}
Lewis P, Perez E, Piktus A, Petroni F, Karpukhin V, Goyal N, K{\"u}ttler H,
  Lewis M, Yih Wt, Rockt{\"a}schel T, et~al. (2020) Retrieval-augmented
  generation for knowledge-intensive nlp tasks. \emph{Advances in neural
  information processing systems} 33:9459--9474.

\bibitem[{Li et~al.(2024)Li, Peng, Wang, \protect\BIBand{} Bai}]{li2024impact}
Li H, Peng J, Wang G, Bai X (2024) The impact of process-vs. outcome-oriented
  reviews on the sales of healthcare services. \emph{Information Systems
  Research} 35(4):1909--1927.

\bibitem[{Li et~al.(2021)Li, Wang, \protect\BIBand{} Wang}]{li2021peer}
Li Z, Wang G, Wang HJ (2021) Peer effects in competitive environments: Field
  experiments on information provision and interventions. \emph{Mis Quarterly}
  45(1):163--191.

\bibitem[{List(2024)}]{list2024field}
List JA (2024) Field experiments: Here today gone tomorrow? \emph{The American
  Economist} 69(2):214--234.

\bibitem[{Liu et~al.(2024)Liu, Wu, Wu, Lu, Chang, \protect\BIBand{}
  Feng}]{liu2024llms}
Liu X, Wu Z, Wu X, Lu P, Chang KW, Feng Y (2024) Are llms capable of data-based
  statistical and causal reasoning? benchmarking advanced quantitative
  reasoning with data. \emph{arXiv preprint arXiv:2402.17644} .

\bibitem[{Lu et~al.(2024)Lu, Lu, Lange, Foerster, Clune, \protect\BIBand{}
  Ha}]{lu2024ai}
Lu C, Lu C, Lange RT, Foerster J, Clune J, Ha D (2024) The ai scientist:
  Towards fully automated open-ended scientific discovery. \emph{arXiv preprint
  arXiv:2408.06292} .

\bibitem[{March(1991)}]{march1991exploration}
March JG (1991) Exploration and exploitation in organizational learning.
  \emph{Organization science} 2(1):71--87.

\bibitem[{Milkman et~al.(2024)Milkman, Ellis, Gromet, Jung, Luscher, Mobarak,
  Paxson, Silvera~Zumaran, Kuan, Berman et~al.}]{milkman2024megastudy}
Milkman KL, Ellis SF, Gromet DM, Jung Y, Luscher AS, Mobarak RS, Paxson MK,
  Silvera~Zumaran RA, Kuan R, Berman R, et~al. (2024) Megastudy shows that
  reminders boost vaccination but adding free rides does not. \emph{Nature}
  631(8019):179--188.

\bibitem[{Milkman et~al.(2022)Milkman, Gandhi, Patel, Graci, Gromet, Ho, Kay,
  Lee, Rothschild, Bogard et~al.}]{milkman2022680}
Milkman KL, Gandhi L, Patel MS, Graci HN, Gromet DM, Ho H, Kay JS, Lee TW,
  Rothschild J, Bogard JE, et~al. (2022) A 680,000-person megastudy of nudges
  to encourage vaccination in pharmacies. \emph{Proceedings of the National
  Academy of Sciences} 119(6):e2115126119.

\bibitem[{Milkman et~al.(2021{\natexlab{a}})Milkman, Gromet, Ho, Kay, Lee,
  Pandiloski, Park, Rai, Bazerman, Beshears et~al.}]{milkman2021megastudies}
Milkman KL, Gromet D, Ho H, Kay JS, Lee TW, Pandiloski P, Park Y, Rai A,
  Bazerman M, Beshears J, et~al. (2021{\natexlab{a}}) Megastudies improve the
  impact of applied behavioural science. \emph{Nature} 600(7889):478--483.

\bibitem[{Milkman et~al.(2021{\natexlab{b}})Milkman, Patel, Gandhi, Graci,
  Gromet, Ho, Kay, Lee, Akinola, Beshears et~al.}]{milkman2021megastudy}
Milkman KL, Patel MS, Gandhi L, Graci HN, Gromet DM, Ho H, Kay JS, Lee TW,
  Akinola M, Beshears J, et~al. (2021{\natexlab{b}}) A megastudy of text-based
  nudges encouraging patients to get vaccinated at an upcoming doctor's
  appointment. \emph{Proceedings of the National Academy of Sciences}
  118(20):e2101165118.

\bibitem[{Nonaka(1994)}]{nonaka1994dynamic}
Nonaka I (1994) A dynamic theory of organizational knowledge creation.
  \emph{Organization science} 5(1):14--37.

\bibitem[{Noy \protect\BIBand{} Zhang(2023)}]{noy2023experimental}
Noy S, Zhang W (2023) Experimental evidence on the productivity effects of
  generative artificial intelligence. \emph{Science} 381(6654):187--192.

\bibitem[{{OpenAI}(2024)}]{openai2024gpt4o}
{OpenAI} (2024) Gpt-4o: Openai's multimodal flagship model.
  \url{https://openai.com/research/gpt-4o}, accessed: October 2025.

\bibitem[{Paley \protect\BIBand{} van~de Ven(2023)}]{paley2023crowdsourcing}
Paley A, van~de Ven N (2023) Crowdsourcing as a tool for creating effective
  nudges: An example for financial oversubscription. \emph{Proceedings of the
  National Academy of Sciences} 120(44):e2308129120.

\bibitem[{Park et~al.(2023)Park, O'Brien, Cai, Morris, Liang, \protect\BIBand{}
  Bernstein}]{park2023generative}
Park JS, O'Brien J, Cai CJ, Morris MR, Liang P, Bernstein MS (2023) Generative
  agents: Interactive simulacra of human behavior. \emph{Proceedings of the
  36th annual acm symposium on user interface software and technology}, 1--22.

\bibitem[{Patil et~al.(2024)Patil, Zhang, Wang, \protect\BIBand{}
  Gonzalez}]{patil2024gorilla}
Patil SG, Zhang T, Wang X, Gonzalez JE (2024) Gorilla: Large language model
  connected with massive apis. \emph{Advances in Neural Information Processing
  Systems} 37:126544--126565.

\bibitem[{Pornpitakpan(2004)}]{pornpitakpan2004persuasiveness}
Pornpitakpan C (2004) The persuasiveness of source credibility: A critical
  review of five decades' evidence. \emph{Journal of applied social psychology}
  34(2):243--281.

\bibitem[{Rowley(2007)}]{rowley2007wisdom}
Rowley J (2007) The wisdom hierarchy: representations of the dikw hierarchy.
  \emph{Journal of information science} 33(2):163--180.

\bibitem[{Saccardo et~al.(2024)Saccardo, Dai, Han, Vangala, Hoo,
  \protect\BIBand{} Fujimoto}]{saccardo2024field}
Saccardo S, Dai H, Han MA, Vangala S, Hoo J, Fujimoto J (2024) Field testing
  the transferability of behavioural science knowledge on promoting
  vaccinations. \emph{Nature Human Behaviour} 8(5):878--890.

\bibitem[{Schick et~al.(2023)Schick, Dwivedi-Yu, Dess{\`\i}, Raileanu, Lomeli,
  Hambro, Zettlemoyer, Cancedda, \protect\BIBand{}
  Scialom}]{schick2023toolformer}
Schick T, Dwivedi-Yu J, Dess{\`\i} R, Raileanu R, Lomeli M, Hambro E,
  Zettlemoyer L, Cancedda N, Scialom T (2023) Toolformer: Language models can
  teach themselves to use tools. \emph{Advances in neural information
  processing systems} 36:68539--68551.

\bibitem[{Schillinger et~al.(2021)Schillinger, Duran, McNamara, Crossley,
  Balyan, \protect\BIBand{} Karter}]{schillinger2021precision}
Schillinger D, Duran ND, McNamara DS, Crossley SA, Balyan R, Karter AJ (2021)
  Precision communication: Physicians' linguistic adaptation to patients'
  health literacy. \emph{Science advances} 7(51):eabj2836.

\bibitem[{Shool et~al.(2025)Shool, Adimi, Saboori~Amleshi, Bitaraf, Golpira,
  \protect\BIBand{} Tara}]{shool2025systematic}
Shool S, Adimi S, Saboori~Amleshi R, Bitaraf E, Golpira R, Tara M (2025) A
  systematic review of large language model (llm) evaluations in clinical
  medicine. \emph{BMC Medical Informatics and Decision Making} 25(1):117.

\bibitem[{So et~al.(2017)So, Kim, \protect\BIBand{} Cohen}]{so2017message}
So J, Kim S, Cohen H (2017) Message fatigue: Conceptual definition,
  operationalization, and correlates. \emph{Communication Monographs}
  84(1):5--29.

\bibitem[{Sun et~al.(2019)Sun, Gao, \protect\BIBand{} Jin}]{sun2019mobile}
Sun T, Gao G, Jin GZ (2019) Mobile messaging for offline group formation in
  prosocial activities: A large field experiment. \emph{Management Science}
  65(6):2717--2736.

\bibitem[{Thakkar et~al.(2016)Thakkar, Kurup, Laba, Santo, Thiagalingam,
  Rodgers, Woodward, Redfern, \protect\BIBand{} Chow}]{thakkar2016mobile}
Thakkar J, Kurup R, Laba TL, Santo K, Thiagalingam A, Rodgers A, Woodward M,
  Redfern J, Chow CK (2016) Mobile telephone text messaging for medication
  adherence in chronic disease: a meta-analysis. \emph{JAMA internal medicine}
  176(3):340--349.

\bibitem[{Thaler \protect\BIBand{} Sunstein(2009)}]{thaler2009nudge}
Thaler RH, Sunstein CR (2009) \emph{Nudge: Improving decisions about health,
  wealth, and happiness} (Penguin).

\bibitem[{Watanabe et~al.(2018)Watanabe, McInnis, \protect\BIBand{}
  Hirsch}]{watanabe2018cost}
Watanabe JH, McInnis T, Hirsch JD (2018) Cost of prescription drug-related
  morbidity and mortality. \emph{Annals of Pharmacotherapy} 52(9):829--837.

\bibitem[{Wu et~al.(2024)Wu, Bansal, Zhang, Wu, Li, Zhu, Jiang, Zhang, Zhang,
  Liu et~al.}]{wu2024autogen}
Wu Q, Bansal G, Zhang J, Wu Y, Li B, Zhu E, Jiang L, Zhang X, Zhang S, Liu J,
  et~al. (2024) Autogen: Enabling next-gen llm applications via multi-agent
  conversations. \emph{First conference on language modeling}.

\bibitem[{Zeigarnik(1938)}]{zeigarnik1938finished}
Zeigarnik B (1938) On finished and unfinished tasks. .

\bibitem[{Zhou et~al.(2023)Zhou, Wang, Yan, \protect\BIBand{}
  Tan}]{zhou2023spoiled}
Zhou T, Wang Y, Yan L, Tan Y (2023) Spoiled for choice? personalized
  recommendation for healthcare decisions: A multiarmed bandit approach.
  \emph{Information Systems Research} 34(4):1493--1512.

\bibitem[{Zhou et~al.(2022)Zhou, Yan, Wang, \protect\BIBand{}
  Tan}]{zhou2022turn}
Zhou T, Yan L, Wang Y, Tan Y (2022) Turn your online weight management from
  zero to hero: A multidimensional, continuous-time evaluation.
  \emph{Management Science} 68(5):3507--3527.

\end{thebibliography}

\clearpage
\setcounter{page}{1}
\begin{APPENDICES}
\raggedbottom  
\clearpage\section{Stage~1 Baseline Message Treatments}\label{app:stage1-messages}

The 13 baseline messages in Stage~1 were designed through a collaborative process between behavioral science experts and large language models, representing the current state of practice for AI-assisted intervention design. Each message operationalizes specific psychological principles from behavioral economics and health communication literature.

\begin{table}[H]
\renewcommand{\arraystretch}{0.78}

\centering
\caption{Stage~1 Message Portfolio: Human-LLM Co-Designed Baseline Treatments}
\label{tab:stage1-messages-appendix}
\footnotesize
\begin{tabular}{@{}p{2.5cm}p{8cm}cp{1.8cm}@{}}
\toprule
\textbf{Name} & \textbf{Message Text} & \textbf{Chars} & \textbf{Strategy} \\
\midrule
default & Hi, it's Dr. Kristen Johnson's office. Review your Rx details here: & 67 & Baseline \\
salience & Hi, it's Dr. Kristen Johnson's office. New prescription details require your review: & 84 & Salience \\
authority & Dr. Kristen Johnson has prepared your prescription details. Review below: & 73 & Authority \\
socialNorms & Dr. Kristen Johnson's office: Most patients find this useful, review your Rx info: & 82 & Social Proof \\
gainFraming & Dr. Kristen Johnson's office: Better health starts with reviewing your Rx below: & 80 & Gain Frame \\
timeliness & Hi, it's Dr. Kristen Johnson's office. While it's fresh, review Rx info below: & 78 & Temporal \\
commitmentPrompt & Dr. Kristen Johnson's office: Ready to review your prescription details? View now: & 82 & Commitment \\
simplification & Hi, it's Dr. Kristen Johnson's office. Review your Rx details here: & 67 & Simplicity \\
emotionalCue & Hi, it's Dr. Kristen Johnson's office. Your health matters - review your Rx: & 76 & Emotion \\
progressFeedback & Dr. Kristen Johnson's office: Final step from your visit - review prescription: & 79 & Progress \\
goalReinforcement & Hi, it's Dr. Kristen Johnson's office. Your wellness journey continues - review Rx: & 83 & Goal \\
futureSelf & Dr. Kristen Johnson's office: Review your Rx — your future self will thank you: & 84 & Future Self \\
socialIdentity & Dr. Kristen Johnson's office: As a valued patient, please review your Rx below: & 79 & Identity \\
\bottomrule
\end{tabular}

\end{table}

\noindent\textbf{Stage~1 Experimental Results.} The randomized controlled trial with 444,691 invitations revealed substantial heterogeneity in message effectiveness, with click-through rates ranging from 54.3\% to 66.1\%. The top three performers were \textit{salience} (66.1\% CTR), \textit{progressFeedback} (64.5\% CTR), and \textit{default} (62.5\% CTR).

\clearpage\section{DIKW Technical Implementation}\label{app:dikw-technical}

This section provides technical details on the DIKW computational architecture for readers interested in implementation specifics. The main text (Section~3.3) presents the agent-unit signatures and a conceptual overview; this appendix provides the complete output-tuple definitions, the bidirectional plan-revision protocol, and execution-mode strategies.

\subsection{Implementation Stack}
The framework is implemented in Python using LangGraph \citep{langgraph2025} for agent orchestration and state management. LangGraph's directed-graph abstraction is used to model the D~$\rightarrow$~I~$\rightarrow$~K~$\rightarrow$~W transitions as nodes, with conditional edges for the bidirectional flow described below (upward output propagation, downward query propagation) and checkpointed graph state for cross-session persistence of the topic repositories. All agent LLM calls invoke Claude-4 (Sonnet) \citep{anthropic2025claude4} through the Anthropic API. The D-Agent and I-Agent additionally have tool access for autonomous code execution: each generates Python code (typically 150--300 lines) that runs in a sandboxed subprocess with \texttt{pandas}, \texttt{scipy}, \texttt{statsmodels}, and \texttt{matplotlib} available; produced artifacts (executable script, CSV statistical tables, structured Markdown report with embedded figures) are persisted under the topic's output handle for downstream agents to consume.

\subsection{DIKW as a Computational Pipeline}

Our multi-agent framework is organized into four hierarchical layers corresponding to the DIKW model: Data, Information, Knowledge, and Wisdom.
Each layer is implemented by a dedicated Agent-Unit that processes structured work units called \textit{topics}.
Topics serve as the fundamental coordination mechanism across the hierarchy, encapsulating both task specifications and their associated processing results.

\noindent\textbf{Topic Notation and Structure}
We denote a topic at level $L \in \{D, I, K, W\}$ with index $i$ as $\mathcal{T}^{(L)}_i$, a structured object containing three components: (i) $\mathcal{T}^{(L)}_i.\text{spec}$, the task specification defining the analytical goal, required inputs, and expected outputs; (ii) $\mathcal{T}^{(L)}_i.\text{output}$, the produced artifacts including code, statistical results, and reports; and (iii) $\mathcal{T}^{(L)}_i.\text{status}$, the execution state indicating whether the topic is pending, in\_progress, completed, or failed.
Topics are organized into level-specific repositories: the Data repository $\mathcal{T}^{(D)} = \{\mathcal{T}^{(D)}_1, \mathcal{T}^{(D)}_2, \ldots\}$ contains all data-level topics, and similarly for Information ($\mathcal{T}^{(I)}$), Knowledge ($\mathcal{T}^{(K)}$), and Wisdom ($\mathcal{T}^{(W)}$).
These repositories enable contextual state management across plan sessions, allowing agents to intelligently reuse previously completed work and selectively regenerate only what is necessary.

With this notation established, we can now describe how Agent-Units transform topics: the Data Agent-Unit $\mathcal{D}$ processes data-level topics $\mathcal{T}^{(D)}_i$, producing outputs $\mathcal{T}^{(D)}_i.\text{output}$; the Information Agent-Unit $\mathcal{I}$ processes $\mathcal{T}^{(I)}_j$; the Knowledge Agent-Unit $\mathcal{K}$ processes $\mathcal{T}^{(K)}_k$; and the Wisdom Agent-Unit $\mathcal{W}$ processes $\mathcal{T}^{(W)}_\ell$.
Topics specify the tasks to be executed, determine dependencies between layers (higher-level topics reference lower-level outputs), and provide the handles for upward output propagation and downward query propagation.

The architecture supports both bottom-up and top-down interactions.
Lower layers (data and information layers) publish outputs upward, enabling higher layers to build progressively richer abstractions, while higher layers (knowledge and wisdom layers) propagate queries downward, triggering the resolution of additional topics when more detailed evidence is required.
This layered architecture supports dynamic adaptation through failure-aware coordination: when a higher-level agent identifies missing dependencies, it propagates queries downward to trigger lower-level topic creation; when a level-specific analysis encounters insurmountable obstacles (e.g., required data missing, computational constraints exceeded), the agent reports diagnostic information upward, initiating plan revision rather than system failure.
This bidirectional communication transforms the traditional feed-forward pipeline into an adaptive knowledge transformation loop, enabling the system to recover from failures and refine its analytical strategy based on encountered obstacles.
In this way, the system decomposes complex tasks into well-defined sub-problems, distributes them across specialized agents, reassembles the results into higher-order insights, and adapts its approach when initial plans prove insufficient.


\subsection{Formal Agent Specifications}

This subsection provides the complete mathematical formalization of the four DIKW agent-units, specifying their inputs, processing operations, and structured outputs.

\subsubsection{Data Agent-Unit Formal Specification}

A \textit{Data Agent-Unit} $\mathcal{D}$ contains specialized internal agents (codewriter, codevalidator, codeexecutor, ReportGenerator) that process data-level topics. The formal transformation is:
\[
\mathcal{D} : \big(D,\, \mathcal{T}^{(D)}_i.\text{spec}\big) \;\longrightarrow\;
\mathcal{T}^{(D)}_i.\text{output} \;=\;
\big(\mathcal{T}^{(D)}_{i,\text{code}},\; \mathcal{T}^{(D)}_{i,\text{report}}\big),
\]
where $D$ denotes the raw dataset, $\mathcal{T}^{(D)}_i.\text{spec}$ specifies the data validation task, $\mathcal{T}^{(D)}_{i,\text{code}}$ contains machine-readable schemas and validators, and $\mathcal{T}^{(D)}_{i,\text{report}}$ provides human-readable summaries of dataset dimensions and quality checks.

\subsubsection{Information Agent-Unit Formal Specification}

An \textit{Information Agent-Unit} $\mathcal{I}$ transforms validated data into individual statistical findings:
\begin{multline*}
\mathcal{I} : \big(D,\, \{\mathcal{T}^{(D)}_i.\text{output}\}_i,\, \mathcal{T}^{(I)}_j.\text{spec}\big) \\
\longrightarrow\ \mathcal{T}^{(I)}_j.\text{output} \;=\; \big(\mathcal{T}^{(I)}_{j,\text{code}},\; \mathcal{T}^{(I)}_{j,\text{report}}\big),
\end{multline*}
where $\{\mathcal{T}^{(D)}_i.\text{output}\}_i$ denotes all available Data-layer artifacts, $\mathcal{T}^{(I)}_{j,\text{code}}$ contains machine-readable statistical measures (test statistics, p-values, confidence intervals), and $\mathcal{T}^{(I)}_{j,\text{report}}$ provides human-readable descriptive statements.

An information-level topic specification consists of four components:
\[
\mathcal{T}^{(I)}_j.\text{spec} = (D',\, C,\, S,\, Q),
\]
where $D'$ is a data slice extracted from raw dataset $D$, $C$ defines the framing context (time windows, units, subgroup definitions), $S$ denotes the variable or group of interest, and $Q$ specifies the descriptive task (compute mean, test correlation, estimate trend).

\subsubsection{Knowledge Agent-Unit Formal Specification}

A \textit{Knowledge Agent-Unit} $\mathcal{K}$ synthesizes multiple Information-level findings into integrated assessments of which principles are reliable:
\[
\mathcal{K}:\ \big(\{\mathcal{T}^{(I)}_j.\text{output}\}_j,\ \mathcal{T}^{(K)}_k.\text{spec}\big)\ \longrightarrow\ \mathcal{T}^{(K)}_k.\text{output},
\]
\begin{multline*}
\mathcal{T}^{(K)}_k.\text{output} = \Big(\mathcal{T}^{(K)}_k.\text{spec},\ r_{\text{theoretical}}, \\
\{\mathcal{T}^{(I)}_{j}.\text{output}\}_{j \in J_k},\ s_{\text{empirical}},\ P\Big),
\end{multline*}
where $r_{\text{theoretical}}$ represents prior or theoretical support, $\{\mathcal{T}^{(I)}_{j}.\text{output}\}_{j \in J_k}$ is the explicit subset of Information outputs synthesized as converging evidence, $s_{\text{empirical}}$ quantifies the consistency and strength of the evidence across multiple Information-level findings, and $P$ records provenance. The index set $J_k$ is determined by the assessment specification: required Information outputs are retrieved if available, otherwise generated dynamically.

\subsubsection{Wisdom Agent-Unit Formal Specification}

A \textit{Wisdom Agent-Unit} $\mathcal{W}$ synthesizes knowledge into actionable solutions:
\begin{multline*}
\mathcal{W}:\ \big(\{\mathcal{T}^{(K)}_k.\text{output}\}_k,\ K^{\text{open}},\ \mathcal{T}^{(W)}_\ell.\text{spec}\big) \\
\longrightarrow\ \mathcal{T}^{(W)}_\ell.\text{output},
\end{multline*}
\begin{multline*}
\mathcal{T}^{(W)}_\ell.\text{output} = \Big(\mathcal{T}^{(W)}_\ell.\text{spec},\ \{\mathcal{T}^{(K)}_k.\text{output}\}_{k \in L_\ell}, \\
K^{\text{open}}_\ell,\ \text{solution},\ P\Big),
\end{multline*}
where $K^{\text{open}}$ represents open-domain knowledge from prior literature or domain principles, $\{\mathcal{T}^{(K)}_k.\text{output}\}_{k \in L_\ell}$ is the subset of knowledge artifacts selected as relevant, $K^{\text{open}}_\ell$ is the external knowledge invoked, $\text{solution}$ represents the proposed strategy, and $P$ records provenance (which knowledge was used, selection rationale, and assumptions applied).


\subsection{Hierarchical Plan-Execute-Revise (HPER) Architecture Details}

This subsection provides detailed technical specifications for the adaptive coordination mechanisms underlying the DIKW framework.

\subsubsection{Plan Sessions and Iterative Refinement}

The framework operates through a sequence of \textit{plan sessions}, each representing one complete attempt to traverse the D→I→K→W pipeline under a governing DIKW Plan. A DIKW Plan specifies the analytical objectives for all four levels: what data should be validated and structured (D), what statistical patterns should be extracted (I), which principles should be assessed for reliability (K), and what interventions should be designed (W). Each session attempts to execute this plan sequentially through the four levels.

When a level completes successfully, meaning all topics at that level have been processed and validated, the orchestrator advances to the next level. However, when a level encounters an insurmountable obstacle, the responsible agent returns a failure report containing: (i) the failed level identifier, (ii) a blocked reason explaining why progress cannot continue, and (iii) suggestions for plan revision. Rather than terminating the entire workflow, this failure triggers plan revision mode.

\subsubsection{Failure Recovery and Plan Revision Protocol}

When an agent reports failure, the system enters plan revision mode, where the orchestrator integrates the failure diagnostics with contextual state information and presents this to human overseers for strategic adjustment. The human reviewer examines the failure report, evaluates whether the suggested plan revision is appropriate, and updates the DIKW Plan accordingly.

Critically, the revised plan does not restart processing from scratch. The system preserves all successfully completed work through contextual state management: validated data artifacts from the Data level, statistical summaries from the Information level, tested hypotheses from the Knowledge level, and generated interventions from the Wisdom level are all retained across plan sessions. When a new session begins under the revised plan, agents intelligently determine which prior work can be reused (enabling efficient resumption) versus which analyses must be regenerated (because objectives or data have changed).

\subsubsection{Execution Mode Selection Strategies}

The framework maintains comprehensive state across plan sessions through accumulated topic repositories: $\mathcal{T}^{(D)}$ (data topics), $\mathcal{T}^{(I)}$ (information topics), $\mathcal{T}^{(K)}$ (knowledge topics), and $\mathcal{T}^{(W)}$ (wisdom topics). Each successfully completed topic (with its task specification, produced artifacts, and execution status) is persisted to these repositories, creating an organizational memory that accumulates across sessions.

This contextual state enables intelligent execution mode selection through four strategies:

\begin{itemize}
\item \textbf{INIT mode}: Loads foundational analyses from predefined templates when starting fresh
\item \textbf{GENERATE mode}: Creates new custom topics via LLM synthesis when addressing novel analytical questions
\item \textbf{LOOP mode}: Intelligently resumes previous work by checking file existence: topics with complete code and reports are SKIPPED (preserving validated results), topics with code but missing reports execute existing scripts (RUN\_CODE mode), and topics without viable code regenerate from scratch
\item \textbf{SKIP mode}: Bypasses an entire level when that level's work is already complete from prior sessions or when strategic revision deems it unnecessary
\end{itemize}

This execution mode intelligence prevents redundant reprocessing (computational efficiency), enables selective refinement of analyses (adaptability), and supports incremental progress across sessions (resilience).

\subsubsection{Progressive Knowledge Accumulation}

The cumulative effect of contextual state management and iterative refinement is that the system functions not as a one-shot pipeline but as a progressive knowledge accumulation engine. Each plan session builds on validated outputs from previous sessions, adding new analyses where needed while preserving successful work. Over multiple sessions, the topic repositories grow to contain comprehensive evidence: diverse data validation checks, rich statistical pattern extractions, integrated assessments of which principles are reliable with supporting evidence chains, and a portfolio of interventions designed under different strategic objectives.

\clearpage\section{DIKW Agent System Prompts}\label{app:agent-prompts}

This appendix presents the task-generation prompts used for each of the four DIKW agent levels. These prompts are taken directly from the system configuration used in the experiment.\footnote{The complete prompt set (28 files covering task generation, aggregation, execution, and orchestration) is available in the code repository.} Template variables (e.g., \texttt{\{data\_instruction\}}) are populated at runtime with project-specific context.

\subsection{Data Agent: Task Generation Prompt}

The Data agent generates tasks for data validation and structural analysis through autonomous code execution.

{\small
\begin{verbatim}
You are tasked with creating data analysis tasks for a research
project. Follow these instructions carefully:

1. First, review the data analysis instruction:
{data_instruction}

2. Examine any human feedback that has been provided to guide
task creation:
{human_task_feedback}

3. Consider the historical context from previous D-steps:
{historical_context}

4. Based on the instruction and context, determine the most
important data analysis themes.

5. Create {max_tasks} specific, actionable data tasks that can
be processed in parallel.

6. For each task:
   - Create a unique, filesystem-safe name with prefix
     "{d_step_prefix}"
   - Provide a clear description of what analysis this task covers
   - Write a detailed step-by-step analysis plan (4-8 steps)
   - Estimate complexity as "simple", "medium", or "complex"

7. Ensure tasks are:
   - Specific and actionable (not vague)
   - Independent enough to run in parallel
   - Comprehensive together (cover the full instruction)
   - Non-overlapping in scope

8. Consider data quality, exploration, modeling, and insights
as potential task areas.
\end{verbatim}
}

\subsection{Information Agent: Task Generation Prompt}

The Information agent generates tasks for statistical pattern extraction through autonomous code execution. Note the explicit distinction between I-level (code execution) and K/W levels (LLM reasoning).

{\small
\begin{verbatim}
You are tasked with creating information extraction tasks that
build upon D-level data analysis results.

IMPORTANT DISTINCTION:
- D-level (Data): Explores raw data structure, types, quality
- I-level (Information): Extracts patterns, statistics,
  correlations FROM the data
- K-level (Knowledge): Synthesizes rules and relationships
  (uses LLM reasoning)
- W-level (Wisdom): Generates strategic recommendations
  (uses LLM reasoning)

I-level tasks should use CODE EXECUTION (run_openhands) to
programmatically extract information. Do NOT use "run_reasoning"
- that is for K/W levels only.

1. Review the I-level instruction:
{information_instruction}

2. Examine the D-level analysis results:
{d_level_context}

3. Consider human feedback (if provided):
{human_task_feedback}

4. Based on D-level findings, create {max_tasks} information
extraction tasks.

5. For each task:
   - Create a unique name with prefix "i{i_step_idx}_"
   - Provide clear description of what information to extract
   - Write detailed step-by-step plan (4-8 specific steps)
   - Set execution_mode to "run_openhands" (code execution)
   - Estimate complexity as "simple", "medium", or "complex"

6. I-level task types to consider:
   - Statistical summaries: descriptive statistics, distributions
   - Correlation analysis: correlation matrices, relationships
   - Pattern identification: trends, cycles, seasonality
   - Anomaly detection: outliers using IQR, z-scores
   - Segment analysis: compare groups, segment-level statistics
   - Distribution analysis: characterize value distributions

7. Each task should:
   - Reference relevant D-level findings as input context
   - Generate code that computes and visualizes the information
   - Produce a markdown report with tables and visualizations
   - Be independent enough to run in parallel

8. Output structure for each task:
   - code/information/{task_name}/main.py (analysis script)
   - report/information/{task_name}/report.md (findings report)
   - report/information/{task_name}/figures/ (visualizations)
\end{verbatim}
}

\subsection{Knowledge Agent: Task Generation Prompt}

The Knowledge agent generates tasks for synthesizing individual Information-level findings into integrated assessments. Note the shift from code execution to LLM reasoning.

{\small
\begin{verbatim}
You are tasked with creating knowledge synthesis tasks from
I-level information extraction results.

IMPORTANT DISTINCTION (DIKW Hierarchy):
- D-level (Data): Explores raw data structure, types, quality
  (uses code execution)
- I-level (Information): Extracts patterns, statistics,
  correlations (uses code execution)
- K-level (Knowledge): Synthesizes rules, principles, causal
  relationships (uses LLM reasoning)
- W-level (Wisdom): Generates strategic recommendations and
  decisions (uses LLM reasoning)

K-level tasks should use LLM REASONING (run_reasoning) to
synthesize knowledge from patterns. This is different from
D/I levels which use code execution.

1. Review the K-level instruction:
{knowledge_instruction}

2. Examine the I-level analysis results (patterns, statistics,
correlations found):
{i_level_context}

3. Consider human feedback (if provided):
{human_task_feedback}

4. Based on I-level patterns, create {max_tasks} knowledge
synthesis tasks.

5. For each task:
   - Create a unique name with prefix "k{k_step_idx}_"
   - Provide clear description of what rules/principles to
     synthesize
   - Write detailed step-by-step plan (4-8 specific steps)
   - Set execution_mode to "run_reasoning"
   - Estimate complexity as "simple", "medium", or "complex"

6. K-level task types to consider:
   - Rule extraction: Business rules, logical relationships
   - Causal analysis: Cause-effect relationships, drivers
   - Relationship mapping: Entity relationships, hierarchies
   - Principle identification: Underlying patterns, invariants,
     generalizations
   - Knowledge structuring: Taxonomies, ontologies,
     conceptual models

7. Each task should:
   - Reference relevant I-level findings as input context
   - Synthesize patterns into actionable rules and principles
   - Produce a markdown report explaining the synthesized
     knowledge
   - Be independent enough to run in parallel

8. Output structure for each task:
   - report/knowledge/{task_name}/report.md
\end{verbatim}
}

\subsection{Wisdom Agent: Task Generation Prompt}

The Wisdom agent generates tasks for translating Knowledge-level assessments into concrete intervention designs.

{\small
\begin{verbatim}
You are tasked with creating wisdom generation tasks from
K-level knowledge synthesis results.

IMPORTANT DISTINCTION (DIKW Hierarchy):
- D-level (Data): Explores raw data structure, types, quality
  (uses code execution)
- I-level (Information): Extracts patterns, statistics,
  correlations (uses code execution)
- K-level (Knowledge): Synthesizes rules, principles, causal
  relationships (uses LLM reasoning)
- W-level (Wisdom): Generates strategic recommendations and
  decisions (uses LLM reasoning)

W-level tasks should use LLM REASONING (run_reasoning) to
generate wisdom from knowledge. This is the highest level of
the DIKW hierarchy - turning knowledge into actionable decisions.

1. Review the W-level instruction:
{wisdom_instruction}

2. Examine the K-level knowledge synthesis results (rules,
principles, relationships found):
{k_level_context}

3. Consider human feedback (if provided):
{human_task_feedback}

4. Based on K-level knowledge, create {max_tasks} wisdom
generation tasks.

5. For each task:
   - Create a unique name with prefix "w{w_step_idx}_"
   - Provide clear description of what recommendations to
     generate
   - Write detailed step-by-step plan (4-8 specific steps)
   - Set execution_mode to "run_reasoning"
   - Estimate complexity as "simple", "medium", or "complex"

6. W-level task types to consider:
   - Strategic recommendations: prioritized actions, roadmaps
   - Risk analysis: potential risks, mitigation strategies
   - Opportunity identification: growth, innovations
   - Decision frameworks: criteria, trade-off analysis
   - Best practices: proven approaches, guidelines
   - Future outlook: predictions, scenarios

7. Each task should:
   - Reference relevant K-level knowledge as input context
   - Transform knowledge into actionable recommendations
   - Produce a markdown report with strategic insights
   - Be independent enough to run in parallel

8. Output structure for each task:
   - report/wisdom/{task_name}/report.md
\end{verbatim}
}

\subsection{LLM Comparison: Pairwise Evaluation Prompt}\label{app:llm-prompts}

The following prompt was used for the frontier LLM comparison reported in Section~4.1. Each LLM (GPT-4o, Claude~3.5 Sonnet) received this system prompt and then evaluated all 380 pairwise comparisons of the 20 Stage~2 message variants.

{\small
\begin{verbatim}
You are an expert in behavioral science, healthcare
communication, and patient engagement.

Your task is to predict which of two prescription review
messages will achieve HIGHER patient engagement and conversion
rates (i.e., more patients clicking to review their
prescription).

Context:
- These are SMS/push notification messages sent to patients
  after a doctor's visit
- Goal: Get patients to click and review their new prescription
  details
- Audience: General patient population across ages and health
  literacy levels
- Messages are from "Dr. Kristen Johnson" (example doctor name)

Your Response:
- Choose which message will likely achieve HIGHER
  engagement/completion rates
- Provide your confidence level (1-10)
- Explain your reasoning briefly (2-3 sentences)

Be objective and base your judgment on behavioral science
principles and healthcare communication best practices.
\end{verbatim}
}

\clearpage\section{Stage~2 Message Templates}\label{app:stage2-messages}

\begin{longtable}{@{}p{0.17\textwidth}p{0.50\textwidth}p{0.08\textwidth}p{0.17\textwidth}@{}}
\caption{Stage~2 Message Templates}
\label{tab:message-templates} \\
\toprule
\textbf{Name} & \textbf{Message} & \textbf{Chars} & \textbf{Generation Strategy} \\
\midrule
\endfirsthead
\toprule
\textbf{Name} & \textbf{Message} & \textbf{Chars} & \textbf{Generation Strategy} \\
\midrule
\endhead
\bottomrule
\endfoot
\bottomrule
\endlastfoot
cognitiveUltra & "Dr. Kristen Johnson: NEW Rx - complete your visit today" & 58 & \multirow{20}{*}{\parbox{2.5cm}{Agentic AI\\Generated}} \\
\textcolor{red}{autonomyMax} & \textcolor{red}{"From Dr. Kristen Johnson: Review your prescription when you're ready"} & \textcolor{red}{69} &  \\
authorityPro & "Dr. Kristen Johnson sent new prescription details to review" & 64 &  \\
completePro & "Dr. Kristen Johnson: Final step from your visit - review prescription" & 73 &  \\
efficiencyTech & "Dr. Kristen Johnson: New Rx info needs quick review" & 54 &  \\
avoidSocial & "Dr. Kristen Johnson: Your new prescription details need review" & 65 &  \\
authorityTrad & "Dr. Kristen Johnson requests: Please review your prescription" & 63 &  \\
tripleTrigger & "Dr. Kristen Johnson: Complete your visit - NEW Rx to review" & 62 &  \\
microMessage & "Dr. Kristen Johnson: New prescription - review" & 47 &  \\
processComplete & "Dr. Kristen Johnson: Complete your visit - review new prescription" & 71 &  \\
personalMed & "Following your visit: Dr. Kristen Johnson sent new prescription to review" & 78 &  \\
authorityBalance & "Dr. Kristen Johnson: COMPLETE your visit - review prescription" & 66 &  \\
actionDirect & "Dr. Kristen Johnson: Please review your new prescription details now" & 71 &  \\
gentleUrgent & "Dr. Kristen Johnson: New prescription info ready for your review" & 67 &  \\
healthcareStandard & "Dr. Kristen Johnson: Review prescription to complete your visit" & 67 &  \\
reciprocityCue & "Dr. Kristen Johnson prepared your prescription - thank you for reviewing" & 75 &  \\
\textcolor{red}{microCommitment} & \textcolor{red}{"Dr. Kristen Johnson's office: Can you review prescription details? Tap below"} & \textcolor{red}{81} &  \\
clarityAction & "Dr. Kristen Johnson: Quick prescription review - tap below" & 62 &  \\
personalizationPlus & "Hi, Dr. Kristen Johnson's office. Your prescription is ready - review today" & 76 &  \\
\textcolor{red}{stepCompletionUrgency} & \textcolor{red}{"Dr. Kristen Johnson: One step left - review your prescription"} & \textcolor{red}{64} &  \\
\midrule
salience & "Hi, it's Dr. Kristen Johnson's office. New prescription details require your review:" & 84 & \multirow{3}{*}{\parbox{2.5cm}{Best from\\Stage~1}} \\
progressFeedback & "Dr. Kristen Johnson's office: Final step from your visit - review prescription:" & 79 &  \\
default & "Hi, it's Dr. Kristen Johnson's office. Please review your prescription below:" & 67 &  \\
\end{longtable}

Table~\ref{tab:message-templates} provides the complete specifications for all 23 message variants (20 newly generated plus 3 from Stage~1), categorized by generation strategy: Agentic AI Generated (all DIKW-generated messages) and Best from Stage~1 (baseline messages carried forward). Messages shown in red (\textcolor{red}{autonomyMax}, \textcolor{red}{microCommitment}, and \textcolor{red}{stepCompletionUrgency}) were omitted from Stage~2 based on the partner's review process, resulting in 20 messages tested.

\clearpage\section{Experimental Design Validation}\label{app:experimental-validation}

%
This appendix presents evidence of proper experimental execution across both stages, covering two diagnostics: (i) cross-sectional randomization balance in arm allocation and (ii) temporal stability of the assignment mechanism across the rollout window.
%
Together these diagnostics underpin the intent-to-treat interpretation of the Stage~2 coefficients in Appendix~\ref{app:regression}.

\noindent\textbf{Randomization Balance}
%
Figures~\ref{fig:r2-message-distribution} and \ref{fig:r3-message-distribution} show the realised allocation of patient invitations across message variants in Stage~1 and Stage~2 respectively.
%
In Stage~1, the 13~variants each received between 33{,}900 and 34{,}700 invitations against an expected per-arm mean of 34{,}207 ($=444{,}691/13$), implying a max-vs-mean deviation below 2.5\%.
%
In Stage~2, the 20~variants each received between 12{,}100 and 12{,}700 invitations against an expected per-arm mean of 12{,}422 ($=248{,}448/20$), implying a max-vs-mean deviation below 5\%.
%
The slightly larger spread in Stage~2 is consistent with the larger number of arms and the shorter rollout window (15~days vs.~18~days), and remains well within the tolerance band expected under independent Bernoulli assignment at these sample sizes.

\begin{figure}[H]
\centering
\includegraphics[width=0.85\textwidth]{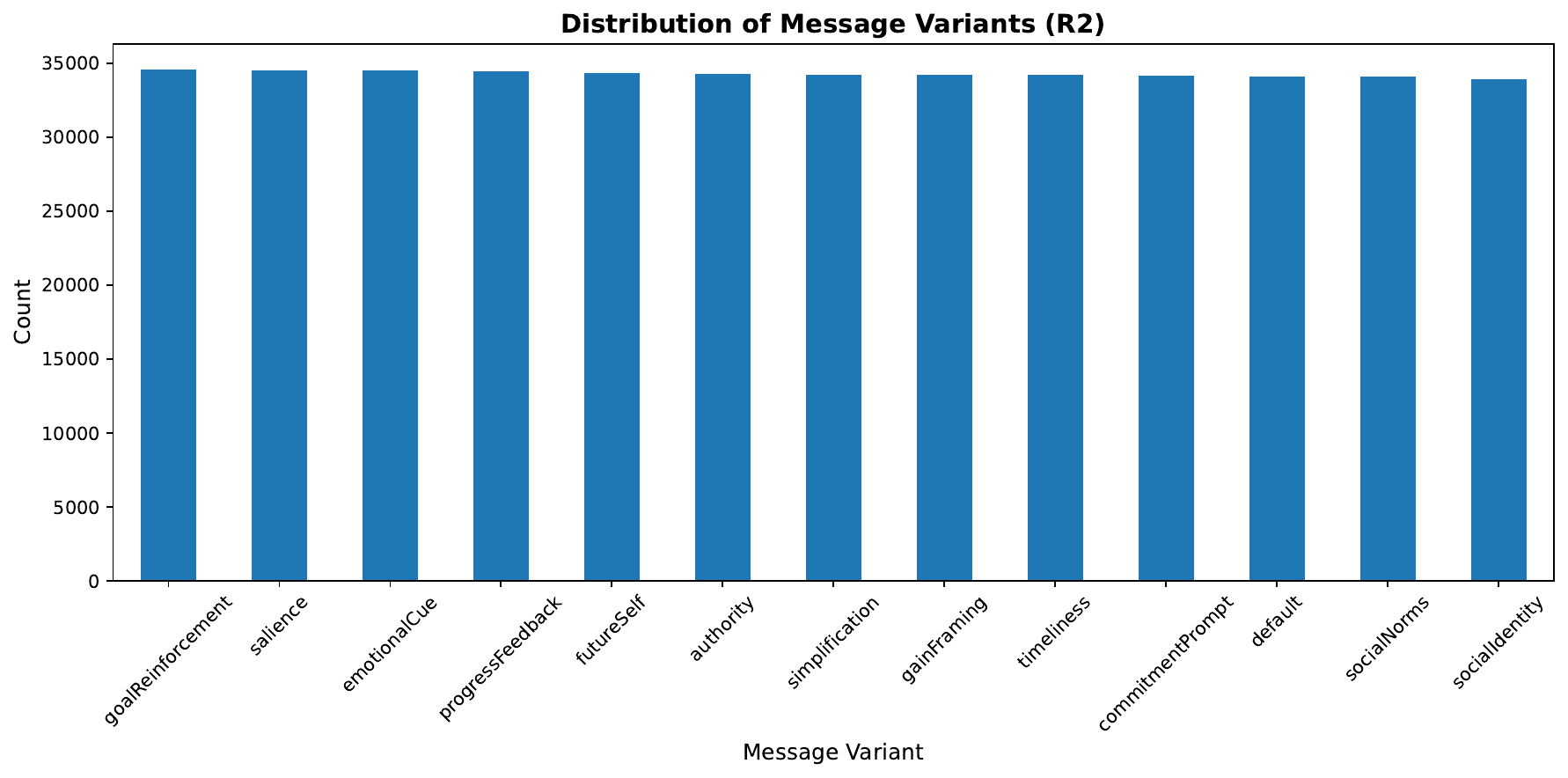}
\caption{Stage~1 Message Variant Allocation (N=444,691)}
\label{fig:r2-message-distribution}
\end{figure}

\begin{figure}[H]
\centering
\includegraphics[width=0.85\textwidth]{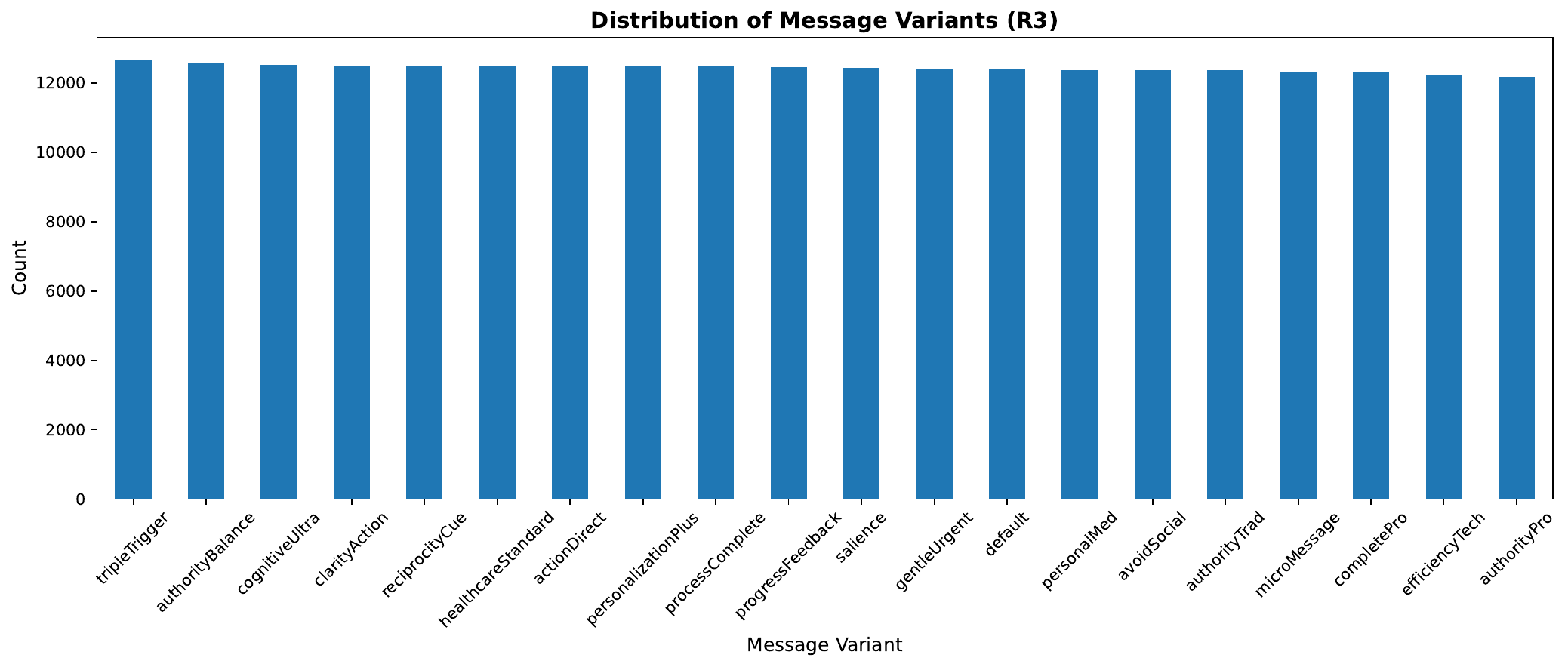}
\caption{Stage~2 Message Variant Allocation (N=248,448)}
\label{fig:r3-message-distribution}
\end{figure}

\noindent\textbf{Temporal Stability of Assignment}
%
Figures~\ref{fig:r2-daily-invitations} and \ref{fig:r3-daily-invitations} decompose total daily invitation volume into per-variant trajectories (top panel) and overall volume (bottom panel).
%
The Stage~1 window (June~16 to July~3, 2025) exhibits a clear weekly cycle: weekday volume peaks near 35{,}000--40{,}000 invitations per day, while the Saturday--Sunday pairs (June~21--22 and June~28--29) drop below 5{,}000, reflecting the underlying clinical prescription pattern rather than any change in the assignment mechanism.
%
The Stage~2 window (August~25 to September~8, 2025) shows the same weekly structure, with an additional dip on Monday September~1 (US Labor Day holiday) that aligns with reduced prescriber activity rather than an experimental anomaly.
%
Critically, the per-variant trajectories in the top panels of both figures track one another in lockstep across all calendar days, including weekend troughs and holiday dips, indicating that the random assignment mechanism applies the same allocation ratios on low-volume days as on high-volume days.
%
This rules out the concern that any single variant is disproportionately exposed to a specific day-of-week or holiday window, which would otherwise bias the temporal-control specification in column~(4) of Appendix~\ref{app:regression}.

\begin{figure}[H]
\centering
\includegraphics[width=0.85\textwidth]{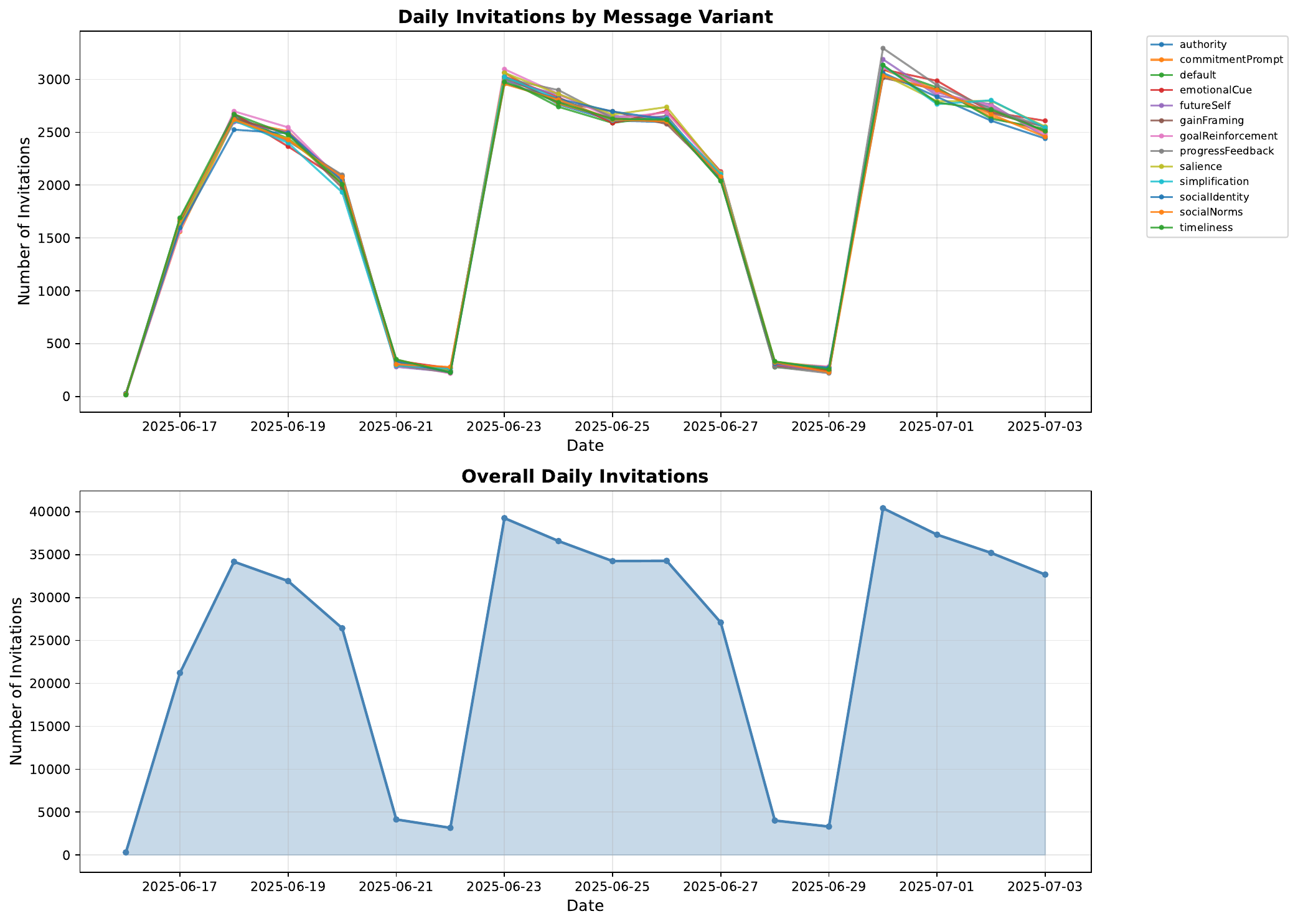}
\caption{Stage~1 Daily Invitation Volume (Jun--Jul 2025)}
\label{fig:r2-daily-invitations}
\end{figure}

\begin{figure}[H]
\centering
\includegraphics[width=0.85\textwidth]{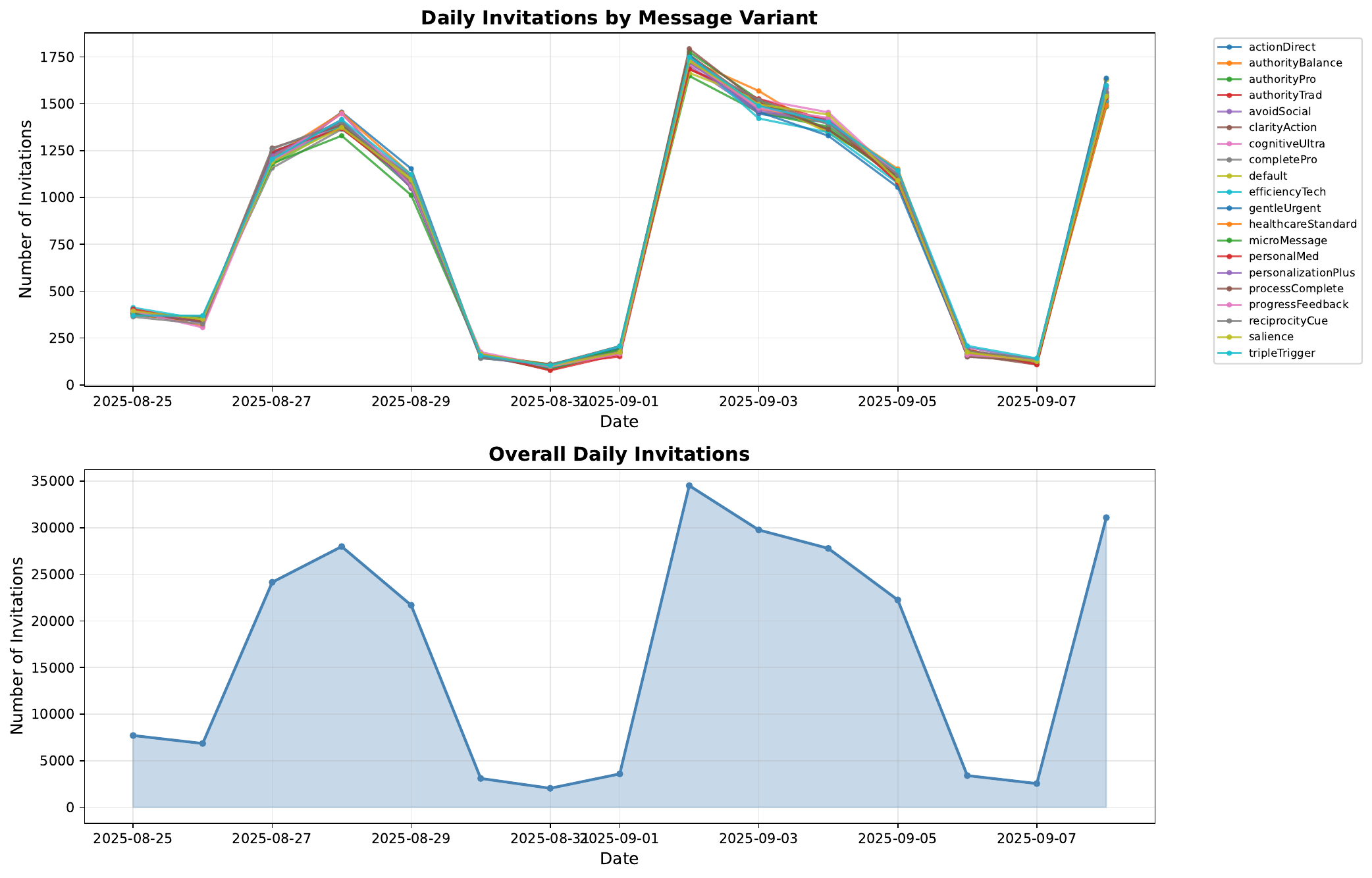}
\caption{Stage~2 Daily Invitation Volume (Aug--Sep 2025)}
\label{fig:r3-daily-invitations}
\end{figure}

\noindent\textbf{Synthesis}
%
Cross-sectional balance (Figures~\ref{fig:r2-message-distribution}--\ref{fig:r3-message-distribution}) and temporal lockstep (Figures~\ref{fig:r2-daily-invitations}--\ref{fig:r3-daily-invitations}) together establish that the experimental allocation is both balanced in aggregate and stationary over the rollout window.
%
This is the empirical foundation for the coefficient-stability result in Appendix~\ref{app:regression} (absolute changes $\leq 0.001$ when demographic, medical, and temporal controls are progressively added), which is the standard signature of a clean randomization.

\clearpage\section{Stage~2 Regression Robustness}\label{app:regression}

\begin{table}[h]

\centering
\caption{Main Treatment Effects on Click-Through Rate — Robustness Across Specifications}
\label{tab:main-effects}
\fontsize{9}{10}\selectfont
\setlength{\tabcolsep}{5pt}
\begin{threeparttable}
\begin{tabular}{l*{7}{c}}
\toprule
\textbf{Message Variant} & \textbf{N} & \mc{\shortstack[t]{Click-\\Through\\Rate (\%)}} & \mc{\shortstack[t]{Authenti-\\cation\\Rate (\%)}} & \mc{\shortstack[t]{(1)\\Simple}} & \mc{\shortstack[t]{(2)\\+Demo}} & \mc{\shortstack[t]{(3)\\+Medical}} & \mc{\shortstack[t]{(4)\\+Temporal}} \\
\midrule
\multicolumn{8}{l}{\textit{\textbf{Agentic AI Generated Messages (17 messages, N=211,173)}}} \\
\quad efficiencyTech         & 12,231 & 69.8 & 59.2 & 0.0654\sym{***} & 0.0645\sym{***} & 0.0645\sym{***} & 0.0644\sym{***} \\
                         &        & & & \mc{(0.0058)} & \mc{(0.0058)} & \mc{(0.0058)} & \mc{(0.0058)} \\
\quad completePro            & 12,305 & 66.5 & 55.1 & 0.0326\sym{***} & 0.0327\sym{***} & 0.0327\sym{***} & 0.0326\sym{***} \\
                         &        & & & \mc{(0.0060)} & \mc{(0.0060)} & \mc{(0.0060)} & \mc{(0.0060)} \\
\quad authorityTrad          & 12,355 & 65.5 & 53.7 & 0.0225\sym{***} & 0.0217\sym{***} & 0.0217\sym{***} & 0.0217\sym{***} \\
                         &        & & & \mc{(0.0060)} & \mc{(0.0060)} & \mc{(0.0060)} & \mc{(0.0060)} \\
\quad clarityAction          & 12,505 & 65.1 & 52.9 & 0.0183\sym{***} & 0.0179\sym{***} & 0.0179\sym{***} & 0.0178\sym{***} \\
                         &        & & & \mc{(0.0060)} & \mc{(0.0060)} & \mc{(0.0060)} & \mc{(0.0060)} \\
\quad healthcareStandard     & 12,493 & 64.8 & 52.7 & 0.0152\sym{**} & 0.0150\sym{**} & 0.0150\sym{**} & 0.0150\sym{**} \\
                         &        & & & \mc{(0.0061)} & \mc{(0.0060)} & \mc{(0.0060)} & \mc{(0.0060)} \\
\quad authorityBalance       & 12,553 & 64.2 & 51.3 & 0.0092 & 0.0081 & 0.0081 & 0.0081 \\
                         &        & & & \mc{(0.0061)} & \mc{(0.0061)} & \mc{(0.0061)} & \mc{(0.0061)} \\
\quad avoidSocial            & 12,357 & 64.1 & 52.3 & 0.0076 & 0.0079 & 0.0079 & 0.0078 \\
                         &        & & & \mc{(0.0061)} & \mc{(0.0061)} & \mc{(0.0061)} & \mc{(0.0061)} \\
\quad tripleTrigger          & 12,673 & 62.0 & 49.3 & -0.0131\sym{**} & -0.0132\sym{**} & -0.0132\sym{**} & -0.0132\sym{**} \\
                         &        & & & \mc{(0.0062)} & \mc{(0.0061)} & \mc{(0.0061)} & \mc{(0.0061)} \\
\quad processComplete        & 12,468 & 61.6 & 48.8 & -0.0167\sym{***} & -0.0164\sym{***} & -0.0164\sym{***} & -0.0166\sym{***} \\
                         &        & & & \mc{(0.0062)} & \mc{(0.0062)} & \mc{(0.0062)} & \mc{(0.0062)} \\
\quad actionDirect           & 12,484 & 61.4 & 48.9 & -0.0186\sym{***} & -0.0177\sym{***} & -0.0177\sym{***} & -0.0177\sym{***} \\
                         &        & & & \mc{(0.0062)} & \mc{(0.0062)} & \mc{(0.0062)} & \mc{(0.0062)} \\
\quad microMessage           & 12,327 & 60.8 & 47.8 & -0.0246\sym{***} & -0.0245\sym{***} & -0.0245\sym{***} & -0.0247\sym{***} \\
                         &        & & & \mc{(0.0062)} & \mc{(0.0062)} & \mc{(0.0062)} & \mc{(0.0062)} \\
\quad gentleUrgent           & 12,412 & 60.8 & 47.4 & -0.0251\sym{***} & -0.0258\sym{***} & -0.0258\sym{***} & -0.0259\sym{***} \\
                         &        & & & \mc{(0.0062)} & \mc{(0.0062)} & \mc{(0.0062)} & \mc{(0.0062)} \\
\quad authorityPro           & 12,164 & 59.9 & 47.1 & -0.0343\sym{***} & -0.0347\sym{***} & -0.0347\sym{***} & -0.0347\sym{***} \\
                         &        & & & \mc{(0.0063)} & \mc{(0.0063)} & \mc{(0.0063)} & \mc{(0.0063)} \\
\quad cognitiveUltra         & 12,521 & 59.6 & 46.4 & -0.0368\sym{***} & -0.0361\sym{***} & -0.0361\sym{***} & -0.0362\sym{***} \\
                         &        & & & \mc{(0.0063)} & \mc{(0.0062)} & \mc{(0.0062)} & \mc{(0.0062)} \\
\quad personalMed            & 12,359 & 57.6 & 43.9 & -0.0567\sym{***} & -0.0565\sym{***} & -0.0565\sym{***} & -0.0564\sym{***} \\
                         &        & & & \mc{(0.0063)} & \mc{(0.0063)} & \mc{(0.0063)} & \mc{(0.0063)} \\
\quad personalizationPlus    & 12,472 & 56.9 & 42.9 & -0.0644\sym{***} & -0.0641\sym{***} & -0.0641\sym{***} & -0.0641\sym{***} \\
                         &        & & & \mc{(0.0063)} & \mc{(0.0063)} & \mc{(0.0063)} & \mc{(0.0063)} \\
\quad reciprocityCue         & 12,494 & 55.5 & 40.9 & -0.0783\sym{***} & -0.0780\sym{***} & -0.0780\sym{***} & -0.0781\sym{***} \\
                         &        & & & \mc{(0.0064)} & \mc{(0.0063)} & \mc{(0.0063)} & \mc{(0.0063)} \\
\\
\multicolumn{8}{l}{\textit{\textbf{BEST FROM STAGE 1 (3 messages, N=37,275)}}} \\
\quad salience               & 12,431 & 66.7 & 55.7 & 0.0337\sym{***} & 0.0340\sym{***} & 0.0340\sym{***} & 0.0339\sym{***} \\
                         &        & & & \mc{(0.0060)} & \mc{(0.0059)} & \mc{(0.0059)} & \mc{(0.0059)} \\
\quad progressFeedback       & 12,459 & 65.5 & 53.2 & 0.0216\sym{***} & 0.0219\sym{***} & 0.0219\sym{***} & 0.0219\sym{***} \\
                         &        & & & \mc{(0.0060)} & \mc{(0.0060)} & \mc{(0.0060)} & \mc{(0.0060)} \\
\quad default (reference)   & 12,385 & 63.3 & 50.3 & \mc{--} & \mc{--} & \mc{--} & \mc{--} \\
                         &        & & & \mc{--} & \mc{--} & \mc{--} & \mc{--} \\
\midrule
Demographics            &        & & & \mc{No}  & \mc{Yes} & \mc{Yes} & \mc{Yes} \\
Medical Context         &        & & & \mc{No}  & \mc{No}  & \mc{Yes} & \mc{Yes} \\
Temporal Controls       &        & & & \mc{No}  & \mc{No}  & \mc{No}  & \mc{Yes} \\
\midrule
Observations            &        & & & \mc{248,448} & \mc{248,448} & \mc{248,448} & \mc{248,448} \\
Pseudo R\sym{2}         &        & & & 0.0040 & 0.0127 & 0.0127 & 0.0129 \\
Baseline CTR            &        & & & 0.6330 & 0.6330 & 0.6330 & 0.6330 \\
\bottomrule
\end{tabular}
\begin{minipage}{\linewidth}
\vspace{0.05cm}
\scriptsize
\textit{Note:} Descriptive columns: per-arm Click-Through Rate and Authentication Rate. Inferential columns (1)--(4): coefficients from logistic regression of click on variant dummies (baseline = \texttt{default}), with robust standard errors in parentheses. FDR-adjusted (Benjamini--Hochberg) $p$-values applied to column (4). \sym{***} $p<0.01$, \sym{**} $p<0.05$, \sym{*} $p<0.10$.
\end{minipage}
\end{threeparttable}

\end{table}

\noindent\textbf{Formal Specification}
%
The four columns in Table~\ref{tab:main-effects} share the base logistic specification:
\begin{equation}
\Pr(\text{click}_i = 1 \mid v_i, X_i) = \Lambda(\alpha + \beta_{v_i} + X_i'\gamma),
\label{eq:logit-appendix}
\end{equation}
where $\Lambda(z) = 1/(1+e^{-z})$, $v_i$ indexes the variant assigned to invitation $i$, $\beta_{\text{default}} = 0$ is the omitted reference, and $X_i$ collects the spec-specific controls.
%
Reported numbers in columns~(1)--(4) are coefficients from logistic regression of click on variant dummies, with robust standard errors in parentheses.
%
The four specifications differ only in $X_i$: column~(1) has $X_i = \emptyset$ (intent-to-treat); column~(2) adds patient age and gender; column~(3) further adds a binary acute/chronic indicator derived from drug therapy class; column~(4) further adds day-of-week and hour-of-day.

\noindent\textbf{Discussion}
%
Table~\ref{tab:main-effects} presents both descriptive engagement rates (per-arm Click-Through Rate and Authentication Rate) and the corresponding regression coefficients on click in columns~(1)--(4).
%
Each inferential column adds successive controls to verify that estimates are stable across specifications, confirming experimental randomization balance.
%
Table~\ref{tab:main-effects-summary} in Section~4.2 highlights the top-performing variants under the same four specifications; the present table reports the full 20-variant breakdown.
%
Coefficient stability across columns (absolute changes $\leq 0.001$ for every variant) validates the randomization protocol reported in Appendix~\ref{app:experimental-validation}, and the coefficient estimates align with the per-arm Click-Through Rate differences shown in the same row.

\clearpage\section{DIKW Agent Outputs and Interpretability}\label{app:agent-outputs}

This appendix presents selected outputs from each layer of the DIKW agent system, illustrating how individual statistical findings are progressively synthesized into integrated assessments and intervention designs. The complete agent outputs are available in the code repository; here we show representative examples and one full interpretability trace.

\subsection{Data Layer Outputs}

The Data agent produces comprehensive metadata documentation about the experimental dataset through autonomous code execution, building an understanding of the data and experimental context without interpretation.

\noindent\textbf{Dataset Characterization.} The agent identifies and documents core structural properties: experimental design with message variant assignments, patient demographic distributions across geographic regions, prescription metadata including therapeutic categories and provider information, and temporal patterns in message delivery schedules. The agent validates data completeness, identifying minimal missing values in core engagement metrics while noting systematic patterns in optional fields such as area deprivation indices.

\noindent\textbf{Experiment Configuration Documentation.} The agent extracts and structures the experimental setup, documenting thirteen distinct message variants with their psychological framing strategies, randomization protocols ensuring balanced assignment across patient demographics, and control group specifications for baseline comparison. This documentation serves as the foundation for all subsequent analytical layers.

\subsection{Information Layer Outputs}

The Information agent transforms validated data into individual statistical findings through autonomous code execution, establishing the empirical foundation for Knowledge-level synthesis.

\noindent\textbf{Engagement Pattern Discovery.} The agent identifies fundamental engagement patterns: click-through rates vary significantly across message variants, with authority-based messages consistently outperforming social proof approaches. Authentication conversion rates remain stable within message strategies but vary across patient demographics. Temporal analysis reveals immediate response preferences, with the majority of engagements occurring within the first hour of message delivery.

\noindent\textbf{Demographic Effect Quantification.} The agent establishes age as the dominant demographic factor in message responsiveness, with engagement increasing progressively across age cohorts. Gender effects prove minimal across all message strategies. Geographic patterns emerge primarily through urban-rural distinctions rather than state-level variations. Medical context analysis reveals that acute conditions drive higher engagement than chronic conditions, while mental health medications show distinct response patterns requiring specialized messaging approaches.

\noindent\textbf{Message Feature Analysis.} Linguistic analysis identifies optimal message length ranges, with concise messages under 65 characters achieving higher engagement. Authority positioning at message opening proves more effective than closing signatures. Action-oriented language consistently outperforms passive informational framing across all patient segments.

\subsection{Knowledge Layer Outputs}

The Knowledge agent synthesizes multiple Information-level findings into integrated assessments of which messaging principles are reliable in the domain, using LLM reasoning rather than code execution.

\noindent\textbf{Principle Reliability Assessment.} The agent synthesizes converging evidence from multiple Information-level analyses: efficiency framing is identified as reliable because message length, action language, and cross-subgroup consistency all point in the same direction. Social proof is identified as ineffective based on consistent underperformance across every demographic and medical subgroup. Completion framing and professional authority are identified as consistent patterns supported by multiple analyses.

\noindent\textbf{Patient Segmentation Strategies.} The agent establishes hierarchical segmentation principles: medical urgency supersedes demographic factors in determining optimal message strategy. Age-based adaptation provides consistent performance improvements across all medical contexts. Condition-specific messaging requirements emerge for mental health, pain management, and cardiovascular medications, each requiring distinct psychological approaches.

\noindent\textbf{Temporal Optimization Patterns.} The agent identifies systematic temporal effects: immediate response windows define engagement success, with exponential decay in response probability after the first hour. Weekday-weekend patterns remain consistent within patient segments but vary across age groups. Time-of-day effects interact with medication types, suggesting circadian influences on health decision-making.


\subsection{Complete Interpretability Trace: Knowledge Claim K1.1}

This subsection demonstrates the DIKW framework's complete interpretability through a detailed trace of Knowledge claim K1.1 (Urgency Dominance in Healthcare Messaging).

\noindent\textbf{Hypothesis Formulation}
The Knowledge Agent-Unit formulated hypothesis K1.1: ``Urgency and salience messaging consistently outperforms social proof messaging across diverse healthcare populations.'' This generalizable claim extends beyond describing dataset patterns to predict performance in future healthcare communication contexts.

\noindent\textbf{Information Retrieval and Integration}
To evaluate this hypothesis, the agent systematically retrieved three Information outputs:

\textit{Information I-2.1 (Message Variant Performance):} Documented that salience achieved 66.06\% click-through rate (n=34,458) compared to socialNorms 54.30\% (n=33,985), a difference of 11.76\%. Statistical testing (I-2.2) confirmed salience significantly outperformed the default baseline (+3.59\%, $t=9.42$, $p<0.001$) while socialNorms significantly underperformed (-8.17\%, $t=-21.18$, $p<0.001$).

\textit{Information I-3.3 (Age-Message Interaction Matrix):} Showed urgency preference was consistent across age groups (15-25: +4.2\%, 35-50: +4.8\%, 50+: +4.1\%) rather than concentrated in specific demographics, strengthening generalizability.

\textit{Information I-5.2 (Medication Type Engagement Patterns):} Demonstrated urgency effectiveness held across acute conditions, chronic conditions, and diverse therapeutic categories, further supporting universal applicability in healthcare contexts.

\noindent\textbf{Confidence Assessment and Theoretical Grounding}
The Knowledge Agent integrated these Information outputs with theoretical support from health psychology literature (healthcare anxiety as constructive channel) and behavioral economics frameworks (present bias in health decisions). The agent assigned confidence score 0.85 based on: strong statistical significance ($p < 0.001$ across multiple tests), large sample size (100,000+ patients), consistent effects across demographic and medical contexts, and alignment with healthcare psychology theory. The agent also documented limitations: effect tested only on SMS channel, may vary with patient-provider relationship strength, and potential cultural context dependencies.

\noindent\textbf{Four-Layer Audit Trail}
This complete reasoning chain provides complete transparency: Raw data (444,691 records) → Information layer (statistical summaries I-2.1, I-2.2, I-3.3, I-5.2) → Knowledge layer (hypothesis K1.1, confidence 0.85) → Wisdom layer (design principles applied to 20 generated messages). This audit trail enables \textit{post hoc} validation, error diagnosis, and systematic refinement.

\end{APPENDICES}

\end{document}